\documentclass[pdflatex,default,iicol]{sn-jnl}

\catcode`\|=12\relax

\usepackage{amssymb}
\usepackage{xspace}  
\usepackage{booktabs}  
\usepackage{subcaption}  
\usepackage{makecell}  
\usepackage{tabularx}  
\usepackage{acro}  
\usepackage{pgfplots}  
\usepackage{xhfill}  
\usepackage[T1]{fontenc}  
\usepackage{pifont}  
\usepackage{placeins}  

\pgfplotsset{compat=1.15}
\usetikzlibrary{calc}
\usetikzlibrary{shadows}
\usetikzlibrary{shadows.blur}
\usetikzlibrary{shapes}

\DeclareAcronym{aad}{short=AAD, long=\textbf{A}dversarial \textbf{A}ccuracy and \textbf{D}ivergence}
\DeclareAcronym{bev}{short=BEV, long=\textbf{B}ird's-\textbf{E}ye \textbf{V}iew}
\DeclareAcronym{bpt}{short=BPT, long=\textbf{B}irthday \textbf{P}aradox \textbf{T}est}
\DeclareAcronym{cd}{short=CD, long=\emph{\textbf{C}hamfer's \textbf{D}istance}}
\DeclareAcronym{cnn}{short=CNN, long=\textbf{C}onvolutional \textbf{N}eural \textbf{N}etwork}
\DeclareAcronym{c2st}{short=C2ST, long=\textbf{C}lassifier Two-\textbf{S}ample \textbf{T}est}
\DeclareAcronym{dnn}{short=DNN, long=\textbf{D}eep \textbf{N}eural \textbf{N}etwork}
\DeclareAcronym{emd}{short=EMD, long=\emph{\textbf{E}arth \textbf{M}over's \textbf{D}istance}}
\DeclareAcronym{fid}{short=FID, long=\emph{\textbf{F}r\'{e}chet \textbf{I}nception \textbf{D}istance}}
\DeclareAcronym{fpd}{short=FPD, long=\emph{\textbf{F}r\'{e}chet \textbf{P}oint Cloud \textbf{D}istance}}
\DeclareAcronym{gam}{short=GAM, long=\textbf{G}enerative \textbf{A}dversarial \textbf{M}etric}
\DeclareAcronym{gan}{short=GAN, long=\textbf{G}enerative \textbf{A}dversarial \textbf{N}etwork}
\DeclareAcronym{is}{short=IS, long=\textbf{I}nception \textbf{S}core}
\DeclareAcronym{irp}{short=IRP, long=\textbf{I}mage \textbf{R}etrieval \textbf{P}erformance}
\DeclareAcronym{lidar}{short=LiDAR, long=\textbf{Li}ght \textbf{D}etection \textbf{a}nd \textbf{R}anging}
\DeclareAcronym{mae}{short=MAE, long=\textbf{M}ean \textbf{A}bsolute \textbf{E}rror}
\DeclareAcronym{mlp}{short=MLP, long=\textbf{M}ulti-\textbf{L}ayer \textbf{P}erceptron}
\DeclareAcronym{mmd}{short=MMD, long=\textbf{M}inimum \textbf{M}atching \textbf{D}istance}
\DeclareAcronym{mos}{short=MOS, long=\textbf{M}ean \textbf{O}pinion \textbf{S}core}
\DeclareAcronym{mse}{short=MSE, long=\textbf{M}ean \textbf{S}quared \textbf{E}rror}
\DeclareAcronym{ndb}{short=NDB, long=\textbf{N}umber of Statistically-\textbf{D}ifferent \textbf{B}ins}
\DeclareAcronym{nrds}{short=NRDS, long=\textbf{N}ormalized \textbf{R}elative \textbf{D}iscriminative \textbf{S}core}
\DeclareAcronym{psnr}{short=PSNR, long=\textbf{P}eak \textbf{S}ignal-to-\textbf{N}oise \textbf{R}atio}
\DeclareAcronym{ssim}{short=SSIM, long=\textbf{S}tructural \textbf{S}imilarity \textbf{I}ndex \textbf{M}easure}
\DeclareAcronym{tsne}{short=t-SNE, long=\textbf{t}-Distributed \textbf{S}tochastic \textbf{N}eighbor \textbf{E}mbedding}
\DeclareAcronym{twrsk}{short=TWRSR, long=\textbf{T}ournament \textbf{W}in \textbf{R}ate and \textbf{S}kill \textbf{R}ating}
\DeclareAcronym{uda}{short=UDA, long=\textbf{U}nsupervised \textbf{D}omain \textbf{A}daptation}

\newcommand{\cmark}{\ding{51}}%
\newcommand{\xmark}{\ding{55}}%

\newcommand{\bp}{\mathbf{p}}
\newcommand{\bS}{\mathbf{S}}
\newcommand{\by}{\mathbf{y}}

\newcommand{\figref}[1]{Fig.~\ref{#1}}
\newcommand{\secref}[1]{Section~\ref{#1}}

\newcommand{\tabref}[1]{Table~\ref{#1}}

\makeatletter
\DeclareRobustCommand\onedot{\futurelet\@let@token\@onedot}
\def\@onedot{\ifx\@let@token.\else.\null\fi\xspace}
\def\eg{e.g\onedot} 
\def\ie{i.e\onedot}

\def\etal{et~al\onedot}

\makeatother

\newcommand{\boldparagraph}[1]{\vspace{0.15cm}\noindent{\bf #1:} }
\newcommand{\boldciteparagraph}[2]{\vspace{0.15cm}\noindent\textbf{#1} #2\textbf{:} }

\definecolor{real_img}{RGB}{215,255,170}
\definecolor{syn_img}{RGB}{19,73,189}
\definecolor{misc_img}{RGB}{144,12,63}
\definecolor{real_plot}{RGB}{100,163,29}
\definecolor{syn_plot}{RGB}{19,73,189}
\definecolor{misc_plot}{RGB}{144,12,63}
\definecolor{cd_plot}{RGB}{27,158,119}
\definecolor{mae_plot}{RGB}{217,95,2}
\definecolor{mse_plot}{RGB}{117,112,179}
\definecolor{tsne_kitti}{RGB}{116,196,118}
\definecolor{tsne_nuscenes}{RGB}{35,139,69}
\definecolor{tsne_pandaset}{RGB}{0,90,50}
\definecolor{tsne_carla}{RGB}{107,174,214}
\definecolor{tsne_builder}{RGB}{33,113,181}
\definecolor{tsne_gtav}{RGB}{8,69,148}
\definecolor{tsne_misc1}{RGB}{189,189,189}
\definecolor{tsne_misc2}{RGB}{150,150,150}
\definecolor{tsne_misc3}{RGB}{115,115,115}
\definecolor{tsne_misc4}{RGB}{82,82,82}
\definecolor{sem_unlabeled}{RGB}{0,0,0}
\definecolor{sem_person}{RGB}{255,30,30}
\definecolor{sem_twowheeler}{RGB}{100,230,245}
\definecolor{sem_largevehicle}{RGB}{0,54,156}
\definecolor{sem_vehicle}{RGB}{100,150,245}
\definecolor{sem_road}{RGB}{255,0,255}
\definecolor{sem_sidewalk}{RGB}{75,0,75}
\definecolor{sem_terrain}{RGB}{150,240,80}
\definecolor{sem_construction}{RGB}{255,200,0}
\definecolor{sem_vegetation}{RGB}{0,175,0}

\definecolor{plot_low}{RGB}{161,215,106}
\definecolor{plot_high}{RGB}{233,163,201}

\newcommand\colorcube[1][black]{\textcolor{#1}{\rule{2.2mm}{2.2mm}}}

\newcommand{\Real}{\textit{Real}}
\newcommand{\Syn}{\textit{Syn}}
\newcommand{\Misc}{\textit{Misc}}

\graphicspath{{./graphics/}}

\jyear{2022}%

\begin{document}

\title[LiDAR Realism Metric]{A Realism Metric for Generated LiDAR Point Clouds}


\author*[1,2]{\fnm{Larissa~T.} \sur{Triess}}\email{larissa.triess@mercedes-benz.com}

\author[1]{\fnm{Christoph~B.} \sur{Rist}}

\author[1,3]{\fnm{David} \sur{Peter}}

\author[2,4]{\fnm{J.~Marius} \sur{Z\"ollner}}

\affil[1]{\orgname{Mercedes-Benz AG}, \orgaddress{\city{Stuttgart}, \country{Germany}}}

\affil[2]{\orgname{Karlsruhe Institute of Technology}, \orgaddress{\city{Karlsruhe}, \country{Germany}}}

\affil[3]{\orgname{Robert-Bosch GmbH}, \orgaddress{\city{Stuttgart}, \country{Germany}}}

\affil[4]{\orgname{Research Center for Information Technology}, \orgaddress{\city{Karlsruhe}, \country{Germany}}}


\abstract{
A considerable amount of research is concerned with the generation of realistic sensor data.
LiDAR point clouds are generated by complex simulations or learned generative models.
The generated data is usually exploited to enable or improve downstream perception algorithms.
Two major questions arise from these procedures:
First, how to evaluate the realism of the generated data?
Second, does more realistic data also lead to better perception performance?

This paper addresses both questions and presents a novel metric to quantify the realism of LiDAR point clouds.
Relevant features are learned from real-world and synthetic point clouds by training on a proxy classification task.
In a series of experiments, we demonstrate the application of our metric to determine the realism of generated LiDAR data and compare the realism estimation of our metric to the performance of a segmentation model.
We confirm that our metric provides an indication for the downstream segmentation performance.
}


\maketitle 

\section{Introduction}
\label{sec:introduction}

Simulations and generative models, such as \acp{gan}, are often used to synthesize realistic training data samples to improve the performance of perception networks~\citep{ParkCVPR2019,Xu2021,Lohdefink2022,Li2022CVPR}.
Assessing the realism of such synthesized samples is a crucial part of the process.
This is usually done by experts, a cumbersome and time consuming approach.
Though a lot of work has been conducted to determine the quality of generated images~\citep{Goodfellow2014NIPS,Salimans2016NIPS,Theis2016ICLR,Heusel2017NIPS,Lehmann2006}, little work is published about how to quantify the realism of point clouds~\citep{Shu2019ICCV,Triess2021GCPR}.
Visual inspection of such data is expensive and not reliable given that the interpretation of 3D point data is rather unnatural for humans.
Because of their subjective nature, it is difficult to compare generative approaches with a qualitative measure.
This work closes the gap and introduces a quantitative evaluation for LiDAR point clouds.

In recent years, a large amount of evaluation measures for \acp{gan} emerged~\citep{Borji2019CVIU}.
Many of them are image-specific and cannot be applied to point clouds.
Existing work on generating realistic LiDAR point clouds mostly relies on qualitative measures to evaluate the generation quality.
Alternatively, some works apply annotation transfer~\citep{Sallab2019ICMLwork} or use the \ac{emd} as an evaluation criterion~\citep{Caccia2019IROS}.
However, these methods require either annotations associated with the data or a matching target, \ie Ground Truth, for the generated sample.
Both are often not feasible when working with large-scale data generation or transfer learning setups.

One main application of data generation is to train downstream perception models, \ie segmentation or detection models that make use of the generated data.
Here it is crucial to reduce the domain gap between generated data and target data on which the trained perception model is applied~\citep{Triess2021IVWork}.
Therefore, the performance of the trained perception model itself can be used as an indication for the realism of the data.
However, using this as a proper metric is impractical since it requires to re-train the target network on multiple versions of the data to evaluate their realism.
A solution is a metric that can determine the realism of the data already while training the generative model.

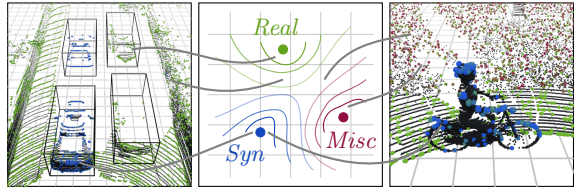
\begin{figure}[t]
    \centering

    \begin{tikzpicture}
        \def\w{(\linewidth / 3) - 0.1cm};

        \node[draw,minimum width=\w,minimum height=\w] (m) at (0, 0) {};
        \node[draw,minimum width=\w,minimum height=\w] (l) at ($(m)-(\linewidth/3,0)$) {};
        \node[draw,minimum width=\w,minimum height=\w] (r) at ($(m)+(\linewidth/3,0)$) {};

        \node[inner sep=0pt] at (l) {\includegraphics[width=\w]{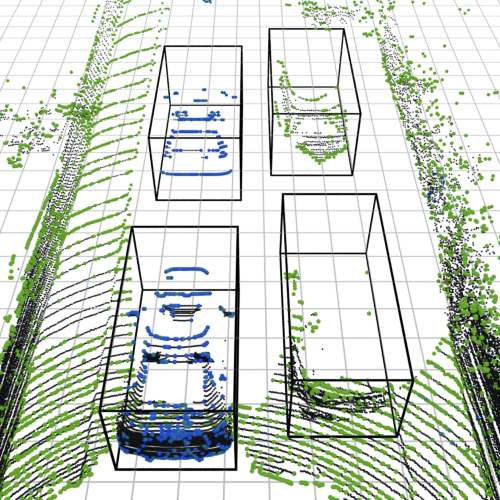}};

        \node[inner sep=0pt] at (r)
        {\includegraphics[width=\w]{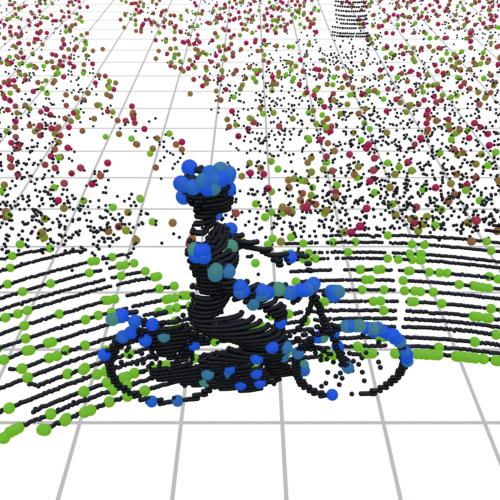}};

        \draw[white!80!black] (-1.1,  0.8) -- (1.1,  0.8);
        \draw[white!80!black] (-1.1,  0.4) -- (1.1,  0.4);
        \draw[white!80!black] (-1.1,  0.0) -- (1.1,  0.0);
        \draw[white!80!black] (-1.1, -0.4) -- (1.1, -0.4);
        \draw[white!80!black] (-1.1, -0.8) -- (1.1, -0.8);
        \draw[white!80!black] ( 0.8, -1.1) -- ( 0.8, 1.1);
        \draw[white!80!black] ( 0.4, -1.1) -- ( 0.4, 1.1);
        \draw[white!80!black] ( 0.0, -1.1) -- ( 0.0, 1.1);
        \draw[white!80!black] (-0.4, -1.1) -- (-0.4, 1.1);
        \draw[white!80!black] (-0.8, -1.1) -- (-0.8, 1.1);
        \path[fill=syn_plot,draw=syn_plot] (-0.4, -0.5) circle (0.06);
        \draw[syn_plot] (-0.6, -0.5) to[out=60,in=190] (-0.3, -0.3) to[out=370,in=110] (-0.2, -0.7);
        \draw[syn_plot,opacity=0.7] (-0.9, -0.55) to[out=40,in=180] (-0.2, -0.2) to[out=360,in=120] (0.0, -0.9);
        \draw[syn_plot,opacity=0.4] (-1.0, -0.3) to[out=20,in=180] (-0.1, 0.0) to[out=360,in=140] (0.3, -1.0);
        \path[fill=real_plot,draw=real_plot] (-0.1, 0.6) circle (0.06);
        \draw[real_plot] (-0.4, 0.7) to[out=290,in=170] (-0.1, 0.4) to[out=350,in=270] (0.2, 0.7);
        \draw[real_plot,opacity=0.7] (-0.8, 0.8) to[out=290,in=170] (-0.2, 0.3) to[out=350,in=270] (0.4, 0.8);
        \draw[real_plot,opacity=0.4] (-1.0, 0.5) to[out=330,in=180] (-0.1, 0.1) to[out=360,in=220] (0.8, 0.9);
        \path[fill=misc_plot,draw=misc_plot] (0.7, -0.3) circle (0.06);
        \draw[misc_plot] (1.0, -0.1) to[out=140,in=20] (0.6, -0.18) to[out=200,in=110] (0.45, -0.5);
        \draw[misc_plot,opacity=0.7] (1.0, 0.2) to[out=220,in=45] (0.5, -0.1) to[out=235,in=130] (0.4, -0.7);
        \draw[misc_plot,opacity=0.4] (1.0, 0.5) to[out=220,in=80] (0.2, -0.4) to[out=260,in=160] (0.6, -1.0);
        \node[text=syn_plot] at (-0.6, -0.8) {\scriptsize\Syn{}\par};
        \node[text=real_plot] at (-0.2, 0.9) {\scriptsize\Real{}\par};
        \node[text=misc_plot] at (0.8, -0.6) {\scriptsize\Misc{}\par};

        \draw[gray,thick] (-2.8, -0.9) to[out=340,in=200] (-0.5, -0.5);
        \draw[gray,thick] (-2.1, 0.6) to[out=10,in=200] (-0.2, 0.5);
        \draw[gray,thick] (-1.6, 0.2) to[out=340, in=200] (-0.1, 0.2);
        \draw[gray,thick] (-0.3, -0.5) to[out=330, in=240] (2.0, -0.4);
        \draw[gray,thick] (0.45, 0.2) to[out=60, in=190] (1.6, 0.8);
        \draw[gray,thick] (0.76, -0.16) to[out=20, in=240] (1.74, 0.4);
    \end{tikzpicture}

    \caption{
    \textbf{Proposed Approach}:
        The realism measure has a tripartite understanding of the 3D-world (middle).
        The left and right image show the color-coded metric scores for query points on two example scenes.
        Both scenes are from the real-world dataset KITTI (\textcolor{real_plot}{\Real{}}) and are augmented with dynamic objects from the simulated CARLA dataset  (\textcolor{syn_plot}{\Syn{}}).
        The left image shows inserted cars from CARLA (left) next to real KITTI cars (right).
        The right image demonstrates the metric results for a synthetic bicycle-and-person object in a KITTI scene.
        Additionally, the terrain in the background is distorted with noise, which is detected as \textcolor{misc_plot}{\Misc{}}.
    }
    \label{fig:eyecatcher}
\end{figure}

To address this need, our previous work~\citep{Triess2021GCPR} proposes a reliable metric that gives a quantitative estimate about the realism of generated LiDAR data.
\figref{fig:eyecatcher}~shows the concept of the metric as a distance measure in high-dimensional feature space.
The metric is trained to learn relevant features via a proxy classification task.
To avoid learning global scene context, we use hierarchical feature set learning to confine features locally in space.
To discourage the network from encoding dataset-specific information, we use an adversarial learning technique which enables robust quantification of unseen data distributions.
In this work, we extend our previous approach~\citep{Triess2021GCPR} with evaluations on the influence of data realism on segmentation performance and add additional ablations of the adversarial training.
In summary, our contributions are:
\begin{itemize}
    \item We present a learning-based quantitative metric to measure the realism of LiDAR point clouds.
    \item We use an adversarial learning technique to suppress irrelevant features, such that the metric can be applied to unseen data.
    \item In experiments on generated LiDAR data, we analyze the relationship between data realism and downstream perception performance.
    We show that our metric is a good indicator for the resulting perception performance.
\end{itemize}

\section{Related Work}
\label{sec:related}

First, this section discusses \ac{gan} evaluation measures and their applicability to generated LiDAR data.
Second, we give a brief overview on metric learning.

\subsection{GAN Evaluation Measures} 
\label{sec:related_gan_measures}

A considerable amount of literature deals with how to evaluate generative models and proposes various evaluation measures.
The most important ones are summarized in extensive survey papers~\citep{Lucic2018NIPS,Xu2018ArXiv,Borji2019CVIU}.
They can be divided into two major categories: qualitative and quantitative measures.

\subsubsection{Qualitative Evaluation} 

Qualitative evaluation~\citep{Goodfellow2014NIPS,Huang2017CVPR,Zhang2017ICCV,Srivastava2017NIPS,Lin2018NIPS,Chen2016NIPS,Mathieu2016NIPS} uses visual inspection of a small collection of examples by humans and is therefore of subjective nature.
It is a simple way to get an initial impression of the performance of a generative model but cannot be performed in an automated fashion.
In other previous work, we use the \acf{mos} testing to verify the realism of generated LiDAR point clouds~\citep{Triess2019IV}.
It was previously introduced in~\citep{Ledig2017CVPR} to provide a qualitative measure for realism in RGB images.
In contrast to~\citep{Ledig2017CVPR}, where untrained people were used to determine the realism, \citep{Triess2019IV}~requires LiDAR experts for the testing process to assure a high enough sensor domain familiarity of the test persons.
This makes the process even more time-consuming and expensive.
Furthermore, the subjective nature of qualitative measures in general makes it difficult to compare performances across different works, even when a large inspection group, such as Mechanical Turk, is used.
Therefore, quantitative metrics are crucial.

\subsubsection{Quantitative Evaluation} 

\begin{table*}
	\centering
	\caption[GAN Evaluation Measures]{
		\textbf{GAN Evaluation Measures}:
		This table categorizes GAN evaluation measures and states their most important pros and cons according to our application.
	}
	\label{tab:metric_gan_measures}

    {\small
	\begin{tabularx}{\linewidth}{@{\hskip 2pt} p{1.8cm} X @{\hskip 6pt} p{4cm} @{\hskip 6pt} p{4cm} @{\hskip 2pt}}
		\toprule
		Category & Metric Examples & $\oplus$ & $\ominus$ \\

		\midrule
		Feature-based
		&
		\acs{is}~\citep{Salimans2016NIPS},
		Modified~\acs{is}~\citep{Gurumurthy2017CVPR},
		Mode Score~\citep{Che2017ICLR},
		AM Score~\citep{Zhou2018ICLR},
		\acs{fid}~\citep{Heusel2017NIPS},
		FPD~\citep{Shu2019ICCV}
		&
		used in many papers with pre-trained models available
		&
		based on features from non-LiDAR datasets (\ie ImageNet~\citep{ImageNet2009} and ShapeNet~\citep{ShapeNet2015})
		\\

		\midrule
		Distribution-based
		&
		Average Log-Likelihood~\citep{Goodfellow2014NIPS,Theis2016ICLR},
		Coverage~\citep{Tolstikhin2017NIPS},
		\acs{mmd}~\citep{Gretton2012JMLR,Achlioptas2018ICLRWORK},
		\acs{bpt}~\citep{Arora2018ICLR},
		\acs{ndb}~\citep{Richardson2018NIPS}
		&
		independent of data modality,
		capture sample diversity and mode collapse
		&
		manual checkpoint selection,
		no absolute measure,
		(additional visual inspection)
		\\

		\midrule
		Classification
		&
		Wasserstein Critic~\citep{Arjovsky2017},
		Classification Performance~\citep{Radford2016,Isola2017CVPR},
		Boundary Distortion~\citep{Santurkar2018ICML},
		\acs{c2st}~\citep{Lehmann2006},
		\acs{aad}~\citep{Yang2017ICLR}
		&
		independent of data modality
		&
		freshly trained discriminators for each test on held-out data,
		no absolute measure
		\\

		\midrule
	    Output Comparison
	    &
	    \acs{irp}~\citep{Wang2016NIPSWORK},
	    Reconstruction Error~\citep{Xiang2017ArXiv}
	    &
	    independent of data modality,
	    per-sample score
	    &
	    high run-time because of nearest neighbor matching
	    \\

	    \midrule
	    Model Comparison
	    &
	    \acs{gam}~\citep{Im2016ArXiv},
	    \acs{twrsk}~\citep{Olsson2018ArXiv},
	    \acs{nrds}~\citep{Zhang2018WACV}
	    &
	    compare different \acs{gan} models against each other
	    &
	    labor intensive, high complexity
	    \\
	    \cline{2-4}
	    &
	    Precision, Recall, F1 Score
	    &
	    simple and fast to compute
	    &
	    only relative performance of discriminator to generator
	    \\

	    \midrule
	    Low-Level Statistics
	    &
	    \acs{ssim}~\citep{Wang2004}, \acs{psnr}, sharpness, contrast, mean power spectrum
	    &
	    simple and fast to compute
	    &
	    specific for camera images,
	    no higher-level information
	    \\

		\bottomrule
	\end{tabularx}
	}

\end{table*}

Quantitative evaluation is performed over a large collection of examples, often in an automated fashion.
\tabref{tab:metric_gan_measures}~categorizes a number of quantitative \ac{gan} measures into six categories according to their properties.

\boldciteparagraph{Feature-based}{\citep{Salimans2016NIPS,Gurumurthy2017CVPR,Heusel2017NIPS,Che2017ICLR,Zhou2018ICLR,Shu2019ICCV}}
Feature-based metrics measure the realism of the data by computing a distance in high-dimensional feature spaces.
The \acf{is}~\citep{Salimans2016NIPS} and the \acf{fid}~\citep{Heusel2017NIPS} are the two most popular metrics and extract their features from the ImageNet dataset~\citep{ImageNet2009}.
This makes them exclusively applicable to camera image data.
The \acf{fpd}~\citep{Shu2019ICCV} is applicable to single-object point clouds, as it is based on features from the PointNet dataset~\citep{Charles2017CVPR}.
In contrast to our method, these measures require labels on the target domain to train the feature extractor, cannot handle variable sized point clouds, and do not provide local scores.
Further, it is only possible to compare a sample to one particular distribution and therefore makes it difficult to obtain a reliable measure on unseen data.

\boldciteparagraph{Distribution-based}{\citep{Goodfellow2014NIPS,Theis2016ICLR,Tolstikhin2017NIPS,Gretton2012JMLR,Achlioptas2018ICLRWORK,Arora2018ICLR,Richardson2018NIPS}}
Most distribution-based measures are independent of the data modality and thus can be used to evaluate \acp{gan}  operating  on  point  clouds.
They successfully capture the sample diversity and mode collapse of the model, but cannot determine the realism of a single sample.
Most approaches are labor intensive as they require manual checkpoint selection and several runs over the test data.

\boldciteparagraph{Classification}{\citep{Arjovsky2017,Radford2016,Isola2017CVPR,Santurkar2018ICML,Lehmann2006,Yang2017ICLR}}
Another common approach is to use classification networks to assess the quality of \ac{gan} outputs.
\Acf{c2st}, for example, assesses whether two samples are drawn from the same distribution.
This requires freshly trained discriminators for each test on a held-out subset of the data.

\boldciteparagraph{Output Comparison}{\citep{Wang2016NIPSWORK,Xiang2017ArXiv}}
Among others, computing reconstruction errors is one common method to assess generated data.
For point clouds, \ac{emd} and \ac{cd} are often used, as they can operate in a permutation-invariant fashion.
These metrics also serve as a basis for some distribution-based measures, such as coverage or \acf{mmd}~\citep{Achlioptas2018ICLRWORK}.
Caccia~\etal~\citep{Caccia2019IROS} use \ac{emd} and \ac{cd} directly as a measure of reconstruction quality on entire scenes captured with a LiDAR scanner.
However, this is only applicable to paired translation \acp{gan} or supervised approaches, because it requires a known target to measure the reconstruction error.

\boldciteparagraph{Model Comparison}{\citep{Im2016ArXiv,Olsson2018ArXiv,Zhang2018WACV}}
There exist two types of model comparison techniques.
The first includes simple metrics that capture the performance of the discriminator relative to the current state of the generator.
The other type focuses on the evaluation of sample diversity and comparison between several \ac{gan} architectures.
However, these measures are labor intensive and of high complexity as they often require several network combinations and trainings.

\boldciteparagraph{Low-Level Statistics}{\citep{Khrulkov2018ICML,Wang2004}}
Computing low-level statistics of the underlying data is easy and fast.
However, statistics like \acf{ssim}, \acf{psnr}, sharpness, or contrast are specific for RGB images and not capable to capture higher-level information.

\vspace{0.15cm}\noindent
This work aims at providing a practical quantitative metric to determine the realism of individual generated samples via learned features.
Therefore, we consider our proposed method as a combination of the following categories: feature-based, distribution-based, and output comparison.

\subsection{Metric Learning} 
\label{sec:related_metric_learning}

The goal of deep metric learning is to learn a feature embedding, such that similar data samples are projected close to each other while dissimilar data samples are projected far away from each other in the high-dimensional feature space.
Common methods use siamese networks trained with contrastive losses to distinguish between similar and dissimilar pairs of samples~\citep{Chicco2021}.
Thereupon, triplet loss architectures train multiple parallel networks with shared weights to achieve the feature embedding~\citep{Hoffer2015SIMBAD,Dong2018ECCV}.
This work uses an adversarial training technique to push features in a similar or dissimilar embedding.

\section{Method}
\label{sec:method}

\begin{figure*}[tb]
	\centering
	\input{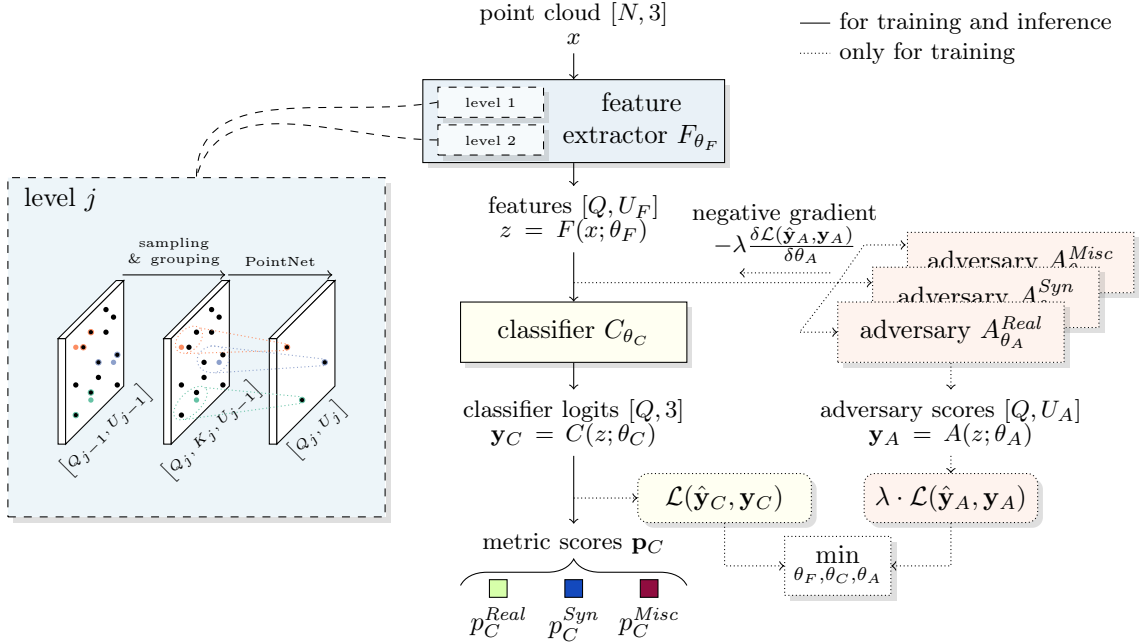}
	\caption{
		\textbf{Architecture}:
		The feature extractor~$F_{\theta_F}$ uses hierarchical feature set learning from PointNet++~\citep{Qi2017NIPS} to encode information about each of the $Q$ query points and their $K$ nearest neighbors.
		The neighborhood features~$z$ are then passed to the classifier~$C_{\theta_C}$ which outputs probability scores~$\bp_C$ for each category (\Real{}, \Syn{}, \Misc{}).
		In training, $z$ is fed to the adversaries~$A_{\theta_A}$, which output probability scores~$\bp_A$ for each dataset of their respective category.
		For the classifier and all three adversaries a multi-class cross-entropy loss is minimized.
		For $C$ to perform as good as possible while $A$ should perform as bad as possible, the gradient is inverted between the adversarial input and the feature extractor~\citep{Beutel2017FAT}.
		$\lambda$~is a factor that regulates the influence of the adversarial loss, weighting the ratio of accuracy versus fairness.
		In our experiments we use a factor of~$\lambda=0.3$.
	}
	\label{fig:architecture}
\end{figure*}

\subsection{Objective and Properties} 
\label{sec:method_properties}

The aim of this work is to provide a method to estimate the level of realism for arbitrary LiDAR point clouds.
We design the metric to learn relevant realism features directly from distributions of real-world data.
The output of the metric can then be interpreted as a distance measure between the input and the learned distribution in a high dimensional space.

Based on the discussed aspects of existing point cloud and \ac{gan} measures, we expect a useful LiDAR point cloud metric to be:

\boldparagraph{Quantitative} 
The realism score is a quantitative measure that determines the distance of the input sample to the internal representation of the learned realistic distribution.
The score~$S^\Real{}$ has well defined lower and upper bounds that reach from $0$ (unrealistic) to $1$ (realistic).

\boldparagraph{Universal} 
The metric has to be applicable to any LiDAR input and therefore must be independent from any application or task.
This means no explicit ground truth information, such as class labels or bounding boxes, is required.

\boldparagraph{Transferable} 
The metric must give a reliable and robust prediction for all inputs, independent of whether the data distribution of the input sample is known by the metric or not.
This makes the metric transferable to new and unseen data.

\boldparagraph{Local} 
The metric should be able to compute spatially local realism scores for smaller regions within a point cloud.
These scores can then be combined with additional information, such as motion, semantics, or distance to provide a detailed analysis of the data.
The metric is also expected to focus on identifying the realism of the point cloud properties while ignoring global scene properties as much as possible to reduce domain biases.

\boldparagraph{Flexible} 
Point clouds are usually sets of un-ordered points with varying size.
Therefore, it is crucial to have a processing that is permutation-invariant and independent of the number of points to process.

\boldparagraph{Simple} 
Easy applicability and a fast computation time allows the metric to run in parallel to the training of a neural network for LiDAR data generation.
This enables monitoring the realism of the generated sample during the training of the network.

\vspace{0.15cm}\noindent
We implement our metric in such a way that the described properties are fulfilled.
To differentiate the metric from a \ac{gan} discriminator, we emphasize that a discriminator is not \textit{transferable} to unseen data, since it recognizes only one specific data distribution to be realistic.

\subsection{Architecture} 
\label{sec:method_architecture}

\figref{fig:architecture}~shows the architecture of our approach.
The following describes the components and presents how each part is designed to contribute towards achieving the desired metric properties.
The underlying idea of the metric design is to compute a distance measure between different data distributions of \textit{realistic} and \textit{unrealistic} LiDAR point cloud compositions.
The network learns features indicating realism from data distributions by using a proxy classification task.
Specifically, the network is trained to classify point clouds from different datasets into three categories: \Real{}, \Syn{}, \Misc{}.
The premise is the possibility to divide the probability space of LiDAR point clouds into those that derive from real-world data (\Real{}), those that derive from simulations (\Syn{}), and all the others (\Misc{}), \eg distorted or randomized data.
Refer to \figref{fig:eyecatcher} for an impression.
By acquiring the prior information about the tripartite data distribution, the metric does not require any target information or labels for inference.

The features are obtained with hierarchical feature set learning, explained in~\secref{sec:method_feature_extractor}.
\secref{sec:method_fairness}~outlines our adversarial learning technique.

\subsubsection{Feature Extractor}
\label{sec:method_feature_extractor}

The blue parts of \figref{fig:architecture} visualize the PointNet++~\citep{Qi2017NIPS} concept of the feature extractor~$F_{\theta_F}$.
It has two abstraction levels, sampling $Q_1\!=\!2048$ and $Q_2\!=\!256$ query points with $K_1\!=\!20$ and $K_2\!=\!10$ nearest neighbors (KNN), respectively.
Keeping the number of neighbors and abstraction levels low limits the network to only encode information about \textit{local} LiDAR-specific statistics instead of global scenery information.
On the other hand, the high amount of query points helps to cover many different regions within the point cloud and guarantees the \textit{local} aspect of our method.
In contrast to PointNet++, we use KNN search instead of radius search to find the neighboring points.
PointNet++ was proposed for point clouds from the ShapeNet dataset~\citep{ShapeNet2015}, which have uniformly sampled points on object surfaces.
In LiDAR point clouds, points are not uniformly distributed and with increasing distance to the sensor, also the distance between neighboring points increase.
Therefore, we found KNN search more practical to obtain meaningful neighborhoods in LiDAR scans compared to radius search.

In each abstraction level, we use a 3-layer MLP with filter sizes of $\left[64, 64, 128\right]$ and $\left[128, 128, 256\right]$, respectively.
This results in the neighborhood features~$z\!=\!F(x,\theta_F)$ of size $\left[Q, U_F\right]$ with $U_F\!=\!256$ features for each of the $Q\!=\!256$ query points.
The features $z$ are then fed to a densely connected classifier~$C_{\theta_C}$ (yellow block).
It consists of a hidden layer with 128 units, to which 50\% dropout is applied during training, and the output layer with $U_C$ units.

The classifier output is a probability vector $\bp_{C,q} =\operatorname{softmax}(y_C) \in [0,1]^{U_C}$ per query point~$q$.
The vector has $U_C\!=\!3$ entries for each of the categories \Real{}, \Syn{} and \Misc{}.
The component $p_{C,q}^\Real{}$ quantifies the degree of realism in each local region~$q$.
The scores $\bS\!=\!\frac{1}{Q} \sum_q \bp_{C,q}$ for the entire scene are given by the mean over all query positions.
Here, $S^\Real{}$ is a measure for the degree of realism of the entire point cloud.
A score of $0$ indicates low realism while $1$ indicates high realism.

\subsubsection{Adversarial Training}
\label{sec:method_fairness}

To obtain a \textit{transferable} metric network, our metric leverages a concept often used to design fair network architectures or domain losses~\citep{Beutel2017FAT,Raff2018DSAA}.
The idea is to force the feature extractor to encode only information into the latent representation~$z$ that is relevant for the realism estimation.
This means, we actively discourage the feature extractor from encoding information that is specific to the distribution of a single dataset.
In other words -- using fair networks terminology~\citep{Beutel2017FAT} -- we treat the concrete dataset name as a sensitive attribute.
With this procedure we can improve the generalization ability towards unknown data.

To achieve this behavior, we add a second output path for adversarial learning that consists of one adversary $A_{\theta_A}$ for each category (see orange parts in~\figref{fig:architecture}).
Each of the adversaries predicts classification probabilities for all the datasets in their respective category.
To simplify the following explanation, we assume there is only one adversary.
The architecture of the adversary is identical to the one of the classifier, except for the number of units in the output layer~$U_A$, which depends on the number of training datasets for the respective category ($U_A^\Real{}=2, U_A^\Syn{}=2, U_A^\Misc{}=3$).
Following the designs proposed in~\citep{Beutel2017FAT,Raff2018DSAA}, we train all network components by minimizing the losses for both heads, $\mathcal{L}_C=\mathcal{L}\left(\by_C,\hat{\by}_C\right)$ and $\mathcal{L}_A=\mathcal{L}\left(\by_A,\hat{\by}_A\right)$, but reversing the gradient in the path between the adversary input and the feature extractor.
The goal is for $C$ to predict the category $\by_C$ and for $A$ to predict the dataset $\by_A$ as good as possible, but for $F$ to make it hard for $A$ to predict $\by_A$.
Training with the reversed gradient results in $F$ encoding as little information as possible for predicting $\by_A$.
The training objective is formulated as
\begin{equation}
\begin{aligned}
    \min_{\theta_F,\theta_C,\theta_A}
	&\mathcal{L} \Big( C\big( F(x;\theta_F);\theta_C \big), \hat{y}_C \Big) \\
	&+
	\mathcal{L} \Big( A\big( J_\lambda[ F(x;\theta_F) ];\theta_A \big), \hat{y}_A \Big)
\end{aligned}
\end{equation}
with $\theta$ being the trainable variables and $J_\lambda$ a special function
\begin{equation}
	J_\lambda[F] = F \quad \text{but} \quad \nabla J_\lambda[F] = -\lambda \cdot \nabla F
\end{equation}
such that the forward pass is an identity function while the gradient is inverted in the backward pass while training.
The factor $\lambda$ determines the ratio of accuracy and fairness.

In the applications of the related literature~\citep{Beutel2017FAT,Raff2018DSAA}, the sensitive attribute and the requested attribute are often correlated but have no direct coupling.
In our case, this would mean that different data samples from the same dataset could belong to multiple categories.
But this is not the case, instead samples from one dataset always belong to the same category.
Therefore, our sensitive attribute, the dataset, always directly determines the requested attribute, the category.
A single adversary would now suppress all information of the sensitive attribute, thus also suppresses important information to obtain the requested attribute which then leads to unwanted decline in classifier performance.
Therefore, a separate adversary for each category is needed, such that only the sensitive information regarding the dataset is suppressed, while keeping the requested information about the category intact.
The adversaries $A:\{A^\Real{}, A^\Syn{}, A^\Misc{}\}$ have the trainable variables $\theta_A:\{\theta_A^\Real{}, \theta_A^\Syn{}, \theta_A^\Misc{}\}$.
Each adversary outputs estimates for only the datasets of their respective category.
This forces the feature extractor to encode only common features within one category, while not removing important features from other categories.
The loss is now defined as $\mathcal{L}_A=\mathcal{L}_{A^\Real{}}+\mathcal{L}_{A^\Syn{}}+\mathcal{L}_{A^\Misc{}}$.

\section{Experimental Setup}
\label{sec:experiments}

\subsection{Datasets} 
\label{sec:setup_datasets}

\begin{table}
	\centering
	\caption{
		\label{tab:datasets}
		\textbf{Datasets}:
		The table lists the datasets for each category.
		The two rightmost columns show whether the dataset is used to train or evaluate the metric model.
		The number of samples used for testing is 1000 for all datasets.
		The number of training samples is listed in the middle column.
	}

	\begin{tabularx}{\linewidth}{lXlcc}
		\toprule  
		 & Dataset & Samples & Train. & Eval. \\
		\midrule  
		\multirow{3}{*}{\rotatebox{90}{\textcolor{real_plot}{\Real{}}}}
		& KITTI~\citep{Geiger2013IJRR} & 18,329 & \cmark & \cmark \\
		& nuScenes~\citep{Caesar2020CVPR} & 28,130 & \cmark & \cmark \\
		& PandaSet~\citep{PandaSet2020} & - & \xmark & \cmark \\
        \midrule  
        \multirow{3}{*}{\rotatebox{90}{\textcolor{syn_plot}{\Syn{}}}}
        & CARLA~\citep{Dosovitskiy2017} & 106,503 & \cmark & \cmark \\
        & GeoSet & 18,200 & \cmark & \cmark \\
        & GTAV-LiDAR~\citep{Hurl2019} & - & \xmark & \cmark \\
        \midrule  
        \multirow{2}{*}{\rotatebox{90}{\textcolor{misc_plot}{\Misc{}}}}
        & Misc~1,2,3 & $\infty$ & \cmark & \cmark \\
        & Misc~4 & - & \xmark & \cmark \\
		\bottomrule  
	\end{tabularx}
\end{table}

\tabref{tab:datasets}~shows the datasets used for this work.
We use two different groups of datasets, one that is used to train and evaluate the metric while the other group is only used for evaluation.
With the strict separation of training and evaluation datasets, additionally to the training and test splits, we demonstrate that our method is a useful measure on unknown data distributions.
In both cases alike, the datasets stem from one of three categories: \Real{}, \Syn{}, \Misc{}.

Within the \Real{} category, publicly available real-world datasets are used for training (KITTI, nuScenes) and evaluation (PandaSet).
For \Syn{}, we use the CARLA simulator where we implement the sensor specifications of a Velodyne HDL-64 sensor to create ray-traced range measurements.
GeoSet is the second dataset in this category.
Here, simple geometric objects, such as spheres and cubes are randomly scattered on a ground plane in three dimensional space and ray-traced in a scan pattern.
Additionally, we augment the synthetic data with little noise at training time, such that they are not trivially distinguishable from the other categories.
For evaluation, we use the GTAV-LiDAR dataset~\citep{Hurl2019}, which contains simulated LiDAR samples from the video game Grand Theft Auto~V~(GTA~V).
It has a large detailed world with realistic graphics, which provides a diverse data collection environment.

Finally, we add a third category, \Misc{}, to allow the network to represent meaningless data distributions, as they often occur during \ac{gan}~trainings or sensor failures.
Therefore, \Misc{} contains randomized data that is generated at training time.
Misc~1 and Misc~2 are generated by linearly increasing the depth over the rows or columns of a virtual LiDAR scanner, respectively.
Misc~3 is a simple Gaussian noise with varying standard deviations.
Misc~4 is only used for evaluation and is created by setting patches of varying height and width of the LiDAR depth projection to the same distance.
Varying degrees of Gaussian noise are added to the Euclidean distances of Misc~\{1,2,4\}.

In addition to the training data listed in the tables, we use 1000 samples from a different split of each dataset to obtain our evaluation results.
No annotations or additional information are required to train or apply the metric, all operations are based on the $xyz$ coordinates of the point clouds.

\subsection{Up-Sampling Models} 
\label{sec:setup_upsampling}

We use the task of up-sampling to demonstrate the application of our metric.
Up-sampling is a type of domain adaptation, where the source domain is the low resolution data and the target domain is the high resolution data.
In contrast to more complex adaptations, such as simulation-to-real or sensor-to-sensor setups, we can focus on evaluating the actual data realism instead of additional domain gaps introduced by scene content.
However, this is still a complex task, since the model must understand the scene in order to synthesize realistic high-resolution LiDAR outputs.
This makes it an ideal testing candidate for our realism metric.

In \secref{sec:application} we compare the realism of generated samples from five different up-sampling methods to the target high-resolution.
The generation process is based on cylindrical depth projections of the LiDAR point clouds, as proposed in~\citep{Triess2019IV}.
We compare two traditional methods, \ie nearest neighbor and bilinear interpolation, and three learning-based methods.
The generator of all three learning-based methods is adapted from the SRGAN architecture~\citep{Ledig2017CVPR}.
One version is trained with an $\mathcal{L}_1$-loss, another with $\mathcal{L}_2$-loss, and the \ac{gan} uses an adversarial loss.
The \ac{gan} discriminator is also adapted from~\citep{Ledig2017CVPR}.
We conduct the experiments for $4\!\times$~up-sampling in the vertical dimension.
Implementation and training details can be found in the appendix.

\subsection{Baselines} 
\label{sec:setup_baseline}

As baselines for our metric, we report the reconstruction errors of the up-sampled data.
These errors can serve as an indication of the generation quality, but are usually not suitable as a metric for synthesized data, since they require a target sample.
In our case, this target is the original high-resolution sample from which we generate the low-resolution sample as input to the up-sampling network.
We compute the \acf{cd}, \ac{mae}, and \ac{mse} between the predicted point cloud~$P^p$ and the target~$P^t$.
For \ac{cd}, the point clouds are considered as un-ordered sets $P=\{p\}$, such that
\begin{equation}
\begin{aligned}
    d_\textit{CD}(P^p,P^t)
    &= \frac{1}{|P^p|} \sum\limits_{p^p \in P^p} \min\limits_{p^t \in P^t} \lVert p^p-p^t \rVert_2 \\
    &+ \frac{1}{|P^t|} \sum\limits_{p^t \in P^t} \min\limits_{p^p \in P^p} \lVert p^t-p^p \rVert_2
\end{aligned}
\end{equation}
while for $\operatorname{MAE}=\lVert p^t_{ij} - p^p_{ij} \rVert_1$ and $\operatorname{MSE}=\lVert p^t_{ij} - p^p_{ij} \rVert_2$, the point clouds are arranged as projected images $P=\{p_{ij}\}$ with the indices $i$ and $j$ for the respective row and column of the projection.
Typical \ac{gan} evaluation measures for point cloud generation are Coverage~\citep{Tolstikhin2017NIPS} and \ac{mmd}~\citep{Gretton2012JMLR}.
Both are based on finding the best match between the generated and the target point cloud.
We can assume that the best match is always the original high-resolution image of the same scene, then the metrics simplify to $\text{Cov}\approx 1.0$ and $\text{MMD}\approx d_\text{CD}$ due to our paired translation.
Therefore, we do not report these metrics additionally to the reconstruction errors in the evaluation section.

\subsection{Semantic Segmentation} 
\label{sec:setup_semseg}

The key application for our metric is to evaluate the generation capabilities of generative models to improve downstream perception.
This enables checkpoint selection or early stopping of \ac{gan} trainings under the assumption that better data leads to better perception models.
We investiagate this in our application experiments.
Using the up-sampling models from \secref{sec:setup_upsampling}, we transform data from the source (low-resolution) to the target (high-resolution) domain.
This step generates pseudo-datasets of different quality for each method.
We then use these pseudo-datasets to train semantic segmentation models which are finally evaluated on the target domain.
It is expected that if the metric ranks the realism of a generated dataset higher than another one, training with this data also leads to better segmentation performance on the target domain.
This is because the data is -- per metric -- more realistic, \ie the domain gap is smaller~\citep{Triess2021IVWork}.

As a segmentation model, we use SqueezeSegV2~\citep{Wu2019ICRA} and RangeNet21~\citep{Milioto2019IROS}.
Instead of the original 19 classes, we combine some of them and only predict 9 classes.
Details on the architecture and training can be found in the appendix.

\section{Metric Evaluation}
\label{sec:evaluation}

\subsection{Balance between Accuracy and Fairness}
\label{sec:experiments_factor}

First, the metric has to be calibrated by choosing the correct factor~$\lambda$ of the adversarial loss during training.
This is an important property which controls the ratio between accuracy and fairness.
A well chosen factor will maximize the difference between a high classifier accuracy and a low adversary accuracy.

\figref{fig:experiments_accuracy_over_factor}~shows the classifier accuracy in black and the adversary accuracy in brown (weighted sum over the three category adversaries, shown as dashed lines).
With increasing~$\lambda$, the adversarial accuracy decreases slowly, while the classification accuracy suddenly drops.
This happens because the classifier gradients are overruled by the reversed gradients of the adversary, hindering it from train properly.
Interestingly, the adversarial part of the \Real{} category is significantly more influenced by~$\lambda$ than those of the other two.
One reason might be that the \Real{} datasets in themselves are already very diverse, especially compared to the \Syn{} or \Misc{} datasets.
The number of different sceneries is higher, but the most variance is caused by more diverse appearance of the same object types (\eg{} pedestrians) and the additional sensor noise, which is not present in the \Syn{} datasets.
This makes it hard for the model to extract only realism relevant features in form of common information from the \Real{} datasets while not removing any other relevant information.
Thus, the model requires more pressure in form of higher~$\lambda$ to accomplish this challenging task for the \Real{} category, compared to \Syn{} and \Misc{}, where it is easier to extract common information while not removing any other relevant information.

We use a factor of $\lambda=0.3$ for all further experiments in this paper (indicated by the gray vertical line).
Here, the classifier has a good performance (93\%) while the adversary operates slightly above chance level (50\%).

\begin{figure}
    \centering

    \begin{tikzpicture}
        \pgfplotsset{
            width=7cm,
            height=4.5cm,
            compat=newest,
            grid style={dashed,gray!30},
        }

        \begin{semilogxaxis}[
            xlabel=Adversary Loss Factor~$\lambda$,
            ylabel=Accuracy,
            xmin=1e-3,
            xmax=1e1,
            ymin=0,
            ymax=1,
            grid=both,
            legend style={at={(0.5,1.1)},anchor=south,nodes={scale=0.8, transform shape}},
            legend columns=2,
            legend cell align={left},
        ]

            \addplot[real_plot,dashed,mark=o,mark options={solid}]
            table[x=factor,y=real,col sep=comma] {data/accuracy_over_factor.csv};
            \addplot[misc_plot,dashed,mark=square,mark options={solid}]
            table[x=factor,y=misc,col sep=comma] {data/accuracy_over_factor.csv};
            \addplot[syn_plot,dashed,mark=triangle,mark options={solid}]
            table[x=factor,y=syn,col sep=comma] {data/accuracy_over_factor.csv};

            \addplot[brown,mark=diamond*] table[x=factor,y=adversary,col sep=comma] {data/accuracy_over_factor.csv};
            \addplot[black,mark=*] table[x=factor,y=classifier,col sep=comma] {data/accuracy_over_factor.csv};

            \addplot[gray,mark=none] coordinates {(4e-1, 0) (4e-1, 1)};

        \legend{Adv.~\Real{},Adv.~\Misc{},Adv.\Syn{},Adversary,Classifier}
        \end{semilogxaxis}
    \end{tikzpicture}

    \caption{
        \textbf{Accuracy vs. Fairness}:
        Accuracy of classifier and adversaries over the loss factor~$\lambda$.
        At small $\lambda$, the classification accuracy is high which means good performance.
        However, adversary accuracy is also quite high (at least for \Real{}) which means no fairness in this part.
        With increasing $\lambda$ the network gets fairer while maintaining its high level of classification accuracy.
        At a certain point the network becomes unstable and deteriorates into chance level performance in the classifier.
    }
    \label{fig:experiments_accuracy_over_factor}
\end{figure}
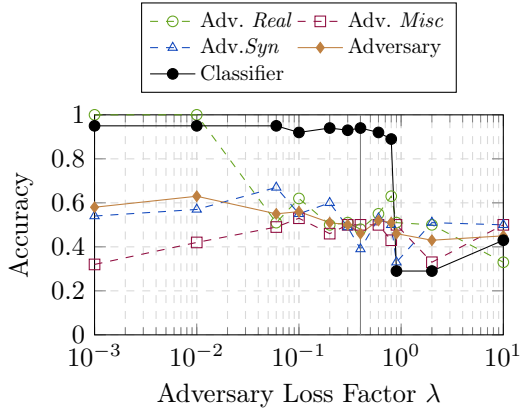

\begin{figure}
    \centering

    {\small
    \begin{tikzpicture}
        \pgfplotsset{
            width=6.6cm,
            compat=newest,
            grid style={dashed,gray!30},
            select coords between index/.style 2 args={
                x filter/.code={
                    \ifnum\coordindex<#1\def\pgfmathresult{}\fi\ifnum\coordindex>#2\def\pgfmathresult{}\fi
                }
            }
        }

        \begin{axis}[
            name=plottop,
            height=2.4cm,
            ylabel=unknown,
            y label style={at={(-0.1,0.5)}},
            ytick={0,1,2},
            yticklabels={
              {\color{misc_plot}Misc~4},
              {\color{syn_plot}GTAV},
              {\color{real_plot}PandaSet}
            },
            yticklabel pos=right,
            xticklabels=\empty,
            xmin=0,
            xmax=1,
            grid=both,
            legend style={at={(0.5,1.2)},anchor=south},
            legend columns=-1,
            legend cell align={left},
        ]

            \addplot[only marks,color=real_plot,mark=*,error bars/.cd,x dir=both,x explicit]
            table[y expr=\coordindex,x=real_mean,x error=real_stddev,col sep=comma,select coords between index={0}{2}]
            {data/metric_training_results.csv};

            \addplot[only marks,color=syn_plot,mark=triangle*,error bars/.cd,x dir=both,x explicit]
            table[y expr=\coordindex,x=syn_mean,x error=syn_stddev,col sep=comma,select coords between index={0}{2}]
            {data/metric_training_results.csv};

            \addplot[only marks,color=misc_plot,mark=square*,error bars/.cd,x dir=both,x explicit]
            table[y expr=\coordindex,x=misc_mean,x error=misc_stddev,col sep=comma,select coords between index={0}{2}]
            {data/metric_training_results.csv};

            \legend{$S^\Real{}$,$S^\Syn{}$,$S^\Misc{}$}
        \end{axis}

        \begin{axis}[
            name=plotbottom,
            at={($(plottop.south)-(0,0.2cm)$)},
            anchor=north,
            height=4.0cm,
            ylabel=known,
            y label style={at={(-0.1,0.5)}},
            xlabel=Mean Softmax Score $S$,
            ytick={3,4,5,6,7,8,9},
            yticklabels={
              {\color{misc_plot}Misc~3},
              {\color{misc_plot}Misc~2},
              {\color{misc_plot}Misc~1},
              {\color{syn_plot}GeomSet},
              {\color{syn_plot}CARLA},
              {\color{real_plot}nuScenes},
              {\color{real_plot}KITTI}
            },
            yticklabel pos=right,
            xmin=0,
            xmax=1,
            grid=both,
        ]

            \addplot[only marks,color=real_plot,mark=*,error bars/.cd,x dir=both,x explicit]
            table[y expr=\coordindex,x=real_mean,x error=real_stddev,col sep=comma,select coords between index={3}{10}]
            {data/metric_training_results.csv};

            \addplot[only marks,color=syn_plot,mark=triangle*,error bars/.cd,x dir=both,x explicit]
            table[y expr=\coordindex,x=syn_mean,x error=syn_stddev,col sep=comma,select coords between index={3}{10}]
            {data/metric_training_results.csv};

            \addplot[only marks,color=misc_plot,mark=square*,error bars/.cd,x dir=both,x explicit]
            table[y expr=\coordindex,x=misc_mean,x error=misc_stddev,col sep=comma,select coords between index={3}{10}]
            {data/metric_training_results.csv};

        \end{axis}
    \end{tikzpicture}
    \par}

    \caption{
        \textbf{Metric Results}:
        Shown is the metric output~$S$ for \Real{}, \Syn{}, and \Misc{} on different datasets.
        The lower part shows the results for the test split of the known datasets, while the upper part depicts one unknown dataset from each category.
        The color of the dataset name indicates the respective category.
    }
    \label{fig:experiments_metric_training_results}
\end{figure}
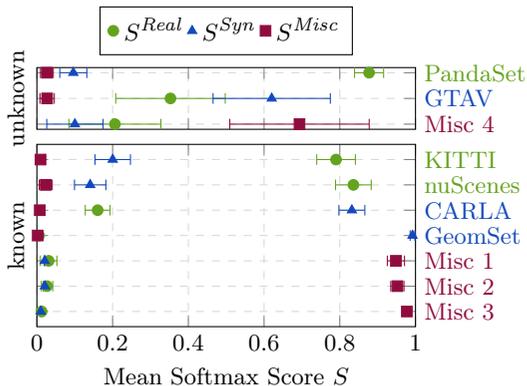

\begin{figure}
    \centering

    \begin{subfigure}[t]{0.31\linewidth}
		\centering
		\frame{\includegraphics[width=\linewidth,angle=180,origin=c]{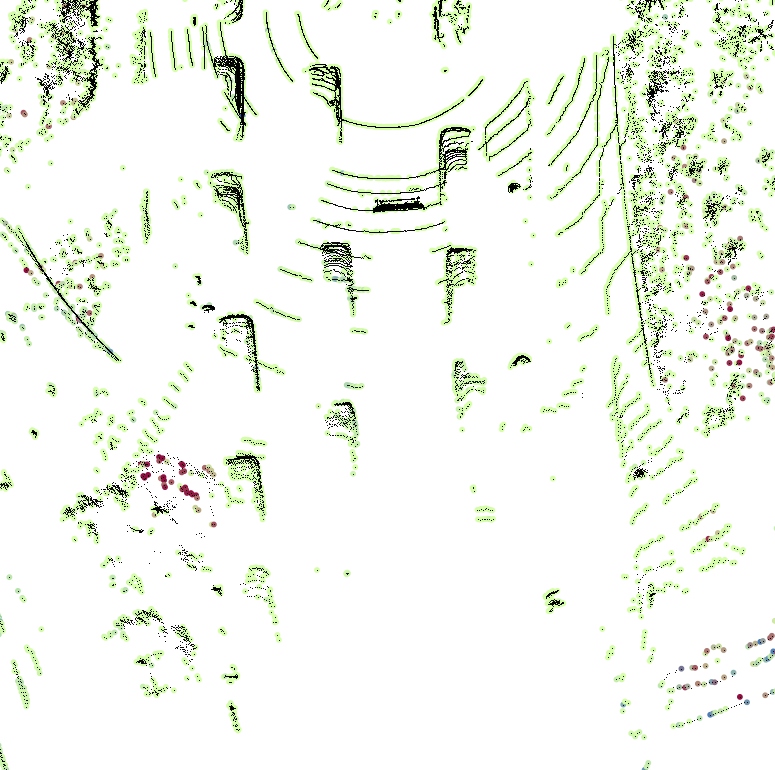}}
		\caption{\label{fig:experiments_unknown_examples_pandaset}PandaSet}
	\end{subfigure}%
    \hspace{4pt}%
    \begin{subfigure}[t]{0.31\linewidth}
		\centering
        \frame{\includegraphics[width=\linewidth,angle=90,origin=c]{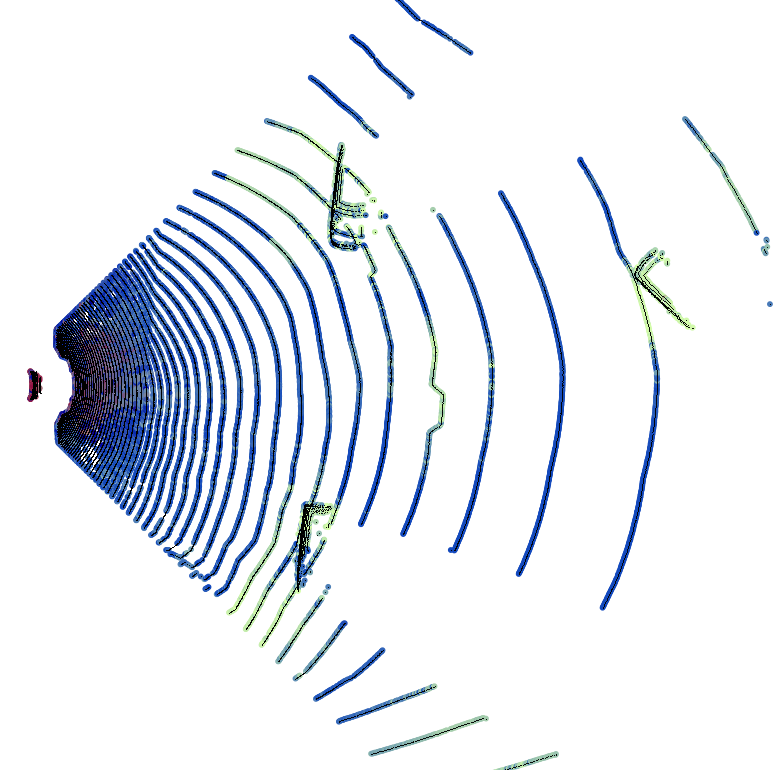}}
		\caption{\label{fig:experiments_unknown_examples_gtav}GTAV}
	\end{subfigure}%
    \hspace{4pt}%
    \begin{subfigure}[t]{0.31\linewidth}
		\centering
        \frame{\includegraphics[width=\linewidth,angle=-90,origin=c]{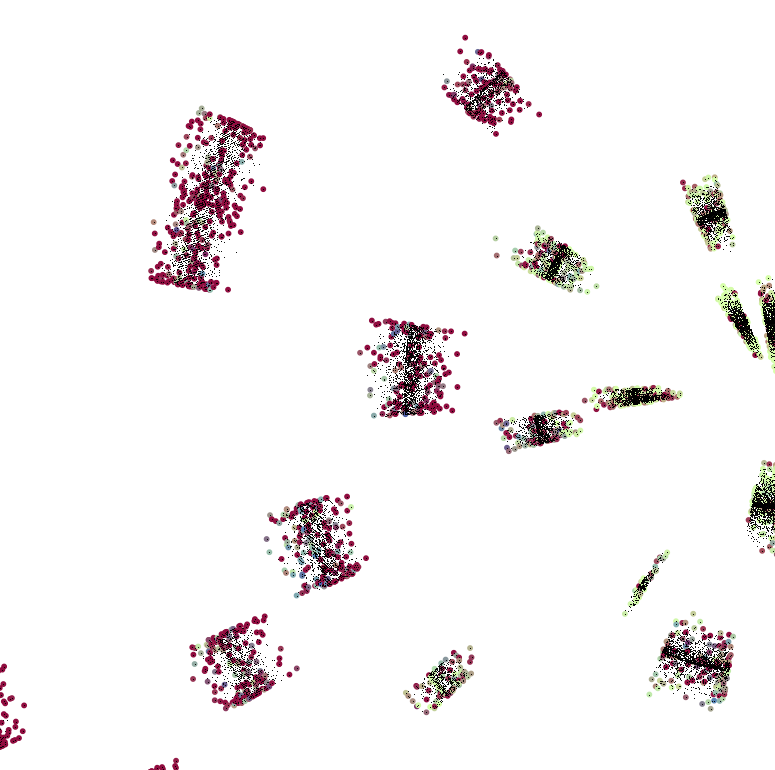}}
		\caption{\label{fig:experiments_unknown_examples_misc4}Misc~4}
	\end{subfigure}%

    \caption{
        \textbf{Qualitative Performance on Unknown Data}:
        The figure shows the metric results on three unknown datasets.
        (\subref{fig:experiments_unknown_examples_pandaset})~shows the PandaSet dataset as an example for \Real{}.
        (\subref{fig:experiments_unknown_examples_gtav})~shows the GTAV dataset for \Syn{}.
        The overall high \Real{} scores seem to be caused by regions that contain cars.
        (\subref{fig:experiments_unknown_examples_misc4})~shows an example for the Misc~4 dataset.
    }
    \label{fig:experiments_unknown_examples}
\end{figure}

\subsection{Overall Dataset Results} 
\label{sec:experiments_classification}

We run our metric network on the evaluation datasets, as well as on the test split of the training datasets.
\figref{fig:experiments_metric_training_results}~shows the mean of the metric scores~$S$ for each of the three categories.
The known datasets (lower part) clearly achieve well-separated scores and predict their respective category, \eg CARLA is classified with a high \Syn{} score.

We obtain notable results on the unknown datasets (upper part).
Qualitative example frames are depicted in~\figref{fig:experiments_unknown_examples}.
The \Real{} dataset PandaSet behaves similar to the two known \Real{} datasets, KITTI and nuScenes.
This shows that the metric focused to encode realism relevant features from KITTI and nuScenes, such that PandaSet is easily categorized as such as well.
The randomly generated Misc~4 dataset is correctly located within the \Misc{} category, however with higher deviations in the scores, leading to \Misc{} scores around 70\% and \Real{} scores around 20\%.
The deviations are caused by the high variance that was used to generate this dataset, where some regions have slightly higher \Real{} or \Syn{} scores.

The \Syn{} dataset GTAV has a slightly different behavior.
Here, $S^\Syn{}$ is around 60\%, while the score for \Real{} is around 35\% and the deviation from those mean values is quite large.
The reason for these high deviations and therefore lower \Syn{} scores is a systematic behavior of the metric caused by the data distribution.
\figref{fig:experiments_unknown_examples_gtav}~shows that the high \Real{} scores mainly stem from regions containing vehicles.
GTAV has more detailed car models than CARLA which therefore appear almost like real vehicles in the point cloud.
This example clearly demonstrates the benefit of the locality aspect of our metric which enables such detailed investigations.

\subsection{Adversary Ablation} 
\label{sec:experiments_fairness}

\begin{figure}
	\centering

    {\scriptsize
    \begin{tabular}{lll}
        \textcolor{tsne_kitti}{\rule{2mm}{2mm}} KITTI & \textcolor{tsne_carla}{\rule{2mm}{2mm}} Carla & \textcolor{tsne_misc1}{\rule{2mm}{2mm}} Misc~1\\
        \textcolor{tsne_nuscenes}{\rule{2mm}{2mm}} nuScenes & \textcolor{tsne_builder}{\rule{2mm}{2mm}} GeoSet & \textcolor{tsne_misc2}{\rule{2mm}{2mm}} Misc~2\\
        & & \textcolor{tsne_misc3}{\rule{2mm}{2mm}} Misc~3\\
        \midrule
        \textcolor{tsne_pandaset}{\rule{2mm}{2mm}} PandaSet & \textcolor{tsne_gtav}{\rule{2mm}{2mm}} GTAV & \textcolor{tsne_misc4}{\rule{2mm}{2mm}} Misc~4\\
    \end{tabular}
    \par}

    \vspace{1em}

	\begin{subfigure}[t]{0.45\linewidth}
		\centering
		\includegraphics[width=\linewidth]{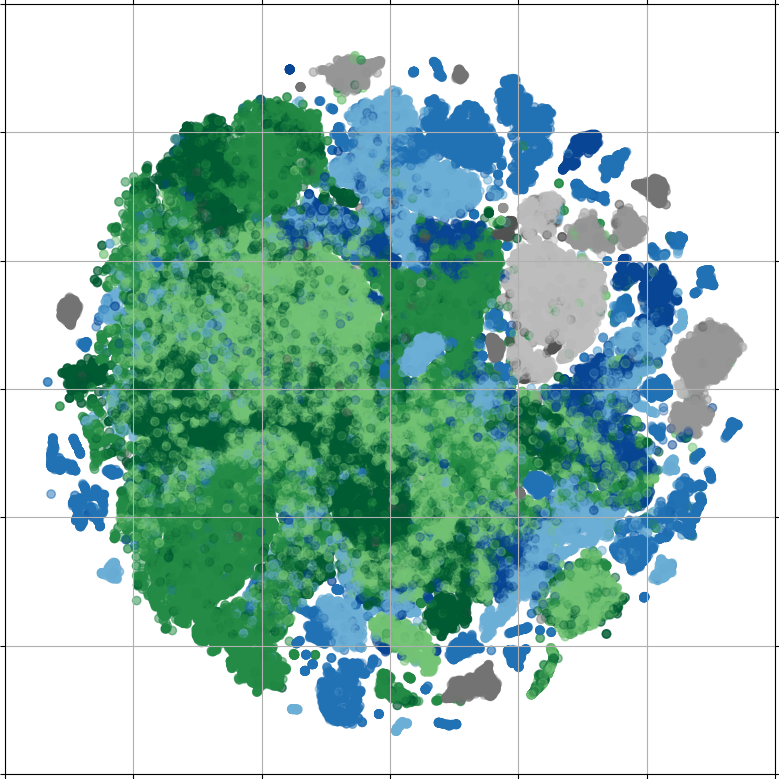}
		\caption{\label{fig:experiments_feature_embedding_none}No adversary}
	\end{subfigure}%
	\hspace{4pt}%
	\begin{subfigure}[t]{0.45\linewidth}
		\centering
		\includegraphics[width=\linewidth]{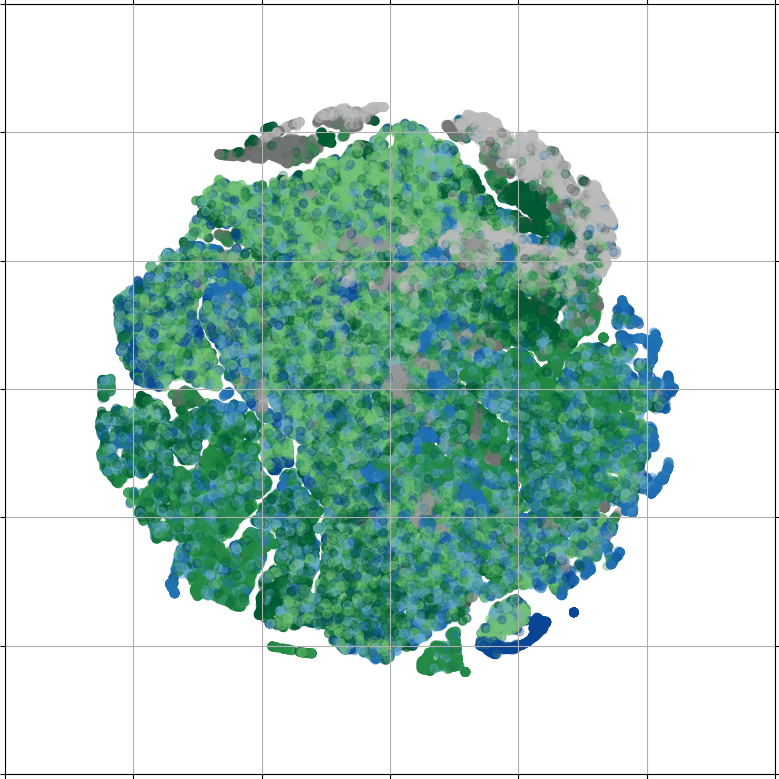}
		\caption{\label{fig:experiments_feature_embedding_full}Common adversary}
	\end{subfigure}

	\vspace{4pt}

    \begin{subfigure}[t]{0.45\linewidth}
        \centering
        \includegraphics[width=\linewidth]{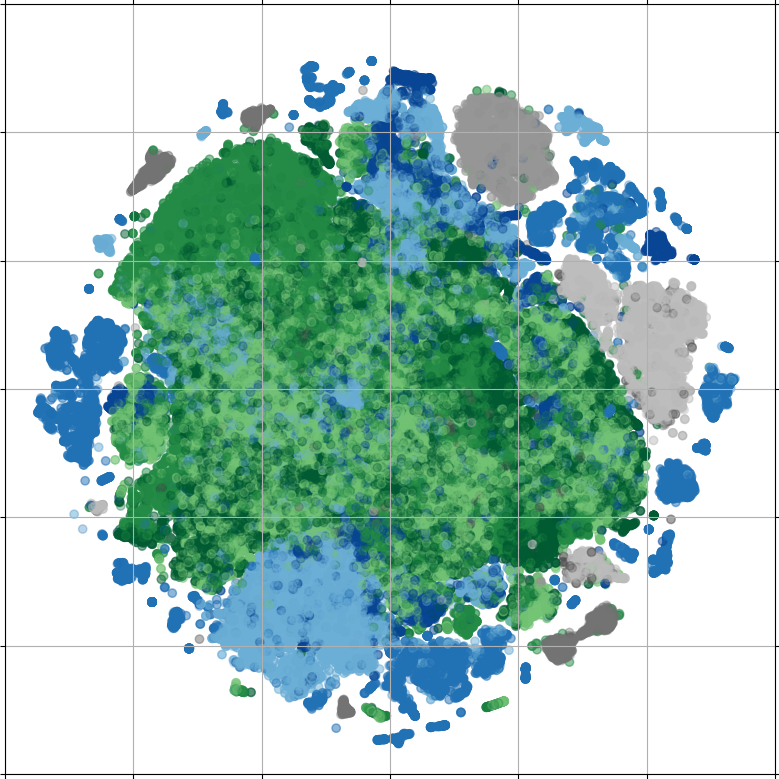}
        \caption{\label{fig:experiments_feature_embedding_real}\Real{} adversary~\citep{Triess2021GCPR}}
    \end{subfigure}%
    \hspace{4pt}%
	\begin{subfigure}[t]{0.45\linewidth}
		\centering
		\includegraphics[width=\linewidth]{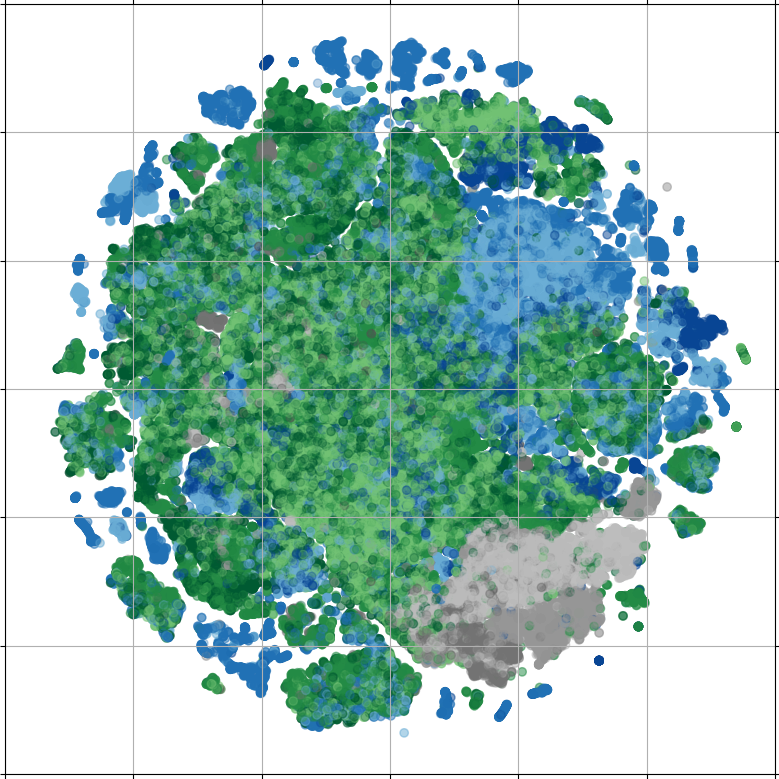}
		\caption{\label{fig:experiments_feature_embedding_category}Individual adversaries [ours]}
	\end{subfigure}

	\caption{
		\textbf{Learned Feature Embedding}:
		Shown are the \ac{tsne} plots for the feature embedding~$z$ of four versions of the adversary configuration for the otherwise identical metric network.
		In (\subref{fig:experiments_feature_embedding_none}) the model is trained without an adversary.
		(\subref{fig:experiments_feature_embedding_full})~shows the features when a single adversary is used for training.
        (\subref{fig:experiments_feature_embedding_real})~visualizes the features of our previous method~\citep{Triess2021GCPR} that only used an adversary for the \Real{} category.
		(\subref{fig:experiments_feature_embedding_category})~depicts our approach, where one adversary per category is trained.
	}

	\label{fig:experiments_feature_embedding}
\end{figure}

The proposed approach uses the adversarial loss to embed features for \Real{}, \Syn{}, and \Misc{} while at the same time omit dataset-specific information as far as possible.
To demonstrate the feature encoding behavior, we train additional metric networks with varying adversary configurations and visualize the learned features on the validation data.

\figref{fig:experiments_feature_embedding}~shows plots of the \ac{tsne} of the neighborhood features~$z$.
\Ac{tsne} is a dimensionality reduction method that tries to map data from a high dimension ($z$ vector) to a low dimension (2D image) space while minimizing information loss.
Close points in the image have similar representations in~$z$.
Each metric category is represented by a different color, while the individual datasets are of different shades of this color.
The darkest colors belong to the unknown datasets that were never seen by the metric network at training time, \ie PandaSet, GTAV, Misc~4.
We include them for demonstration purposes regarding the \emph{transferability} to unseen data.

The two extreme cases of the configuration form \figref{fig:experiments_feature_embedding_none} and \figref{fig:experiments_feature_embedding_full}.
\figref{fig:experiments_feature_embedding_none}~represents the metric as a simple classifier without an adversary, where each shade of each color forms their own clusters with little overlap to others.
This means the features of each dataset are distinct and make it hard for the metric to estimate a reasonable score for unseen datasets.
\figref{fig:experiments_feature_embedding_full}, on the other hand, uses one common adversary which leads to decreased classifier accuracy since features from all sources are forced into a common representation.
This can be observed by the mixed colors with no clusters, not even between categories.

A useful metric requires a mix of the two versions above, where features of one category are similar and features from different categories are dissimilar.
Therefore, we propose to use per-category adversaries.
In our previous work~\citep{Triess2021GCPR} the adversary was only applied for \Real{}, as depicted in \figref{fig:experiments_feature_embedding_real}.
In this work we use one adversary for each category, as represented by \figref{fig:experiments_feature_embedding_category}.
In both cases the green colors of the \Real{} datasets are clearly mixed, while at the same time being sufficiently distinguishable from the blue or gray clusters.
However, our per-category approach (\figref{fig:experiments_feature_embedding_category}) also shows mixed features among the blue and gray points, whereas our previous approach shows more distinct clusters.
This is especially visible for \Misc{}, where \figref{fig:experiments_feature_embedding_real} has one cluster for each shade but our method better combines them.

Further, the feature visualization shows that the unknown dataset PandaSet is fully integrated into the \Real{} cluster for our method, as opposed to when using no adversary.
The clusters of the unknown GTAV dataset mostly overlap with \Syn{}, but also partially with \Real{}.
This aligns with the metric results that we saw previously for GTAV, where parts of the data containing vehicles appear quite realistic.

We conduct the adversary ablation only qualitatively, because it is not possible to compare the quantitative scores of the different versions.
A metric trained as in \figref{fig:experiments_feature_embedding_full} could have a different allocation of scores in range $[0,1]$ than a metric as in \figref{fig:experiments_feature_embedding_category}.

\begin{figure*}
    \centering

    \begin{tikzpicture}
        \pgfplotsset{
            compat=newest,
            grid style={dashed,gray!30},
        }

        {\small

        \begin{axis}[
            name=plot1,
            height=4cm,
            width=5.2cm,
            ytick={0,1,2,3,4,5},
            yticklabels={KITTI,$\mathcal{L}_1$-CNN,Bilinear,$\mathcal{L}_2$-CNN,GAN,Nearest},
            xlabel=Reconstruction Error,
            xmin=0,
            xmax=3.6,
            grid=both,
            legend style={at={(0.5,1.02)},anchor=south,draw=none,nodes={scale=0.8, transform shape}},
            legend columns=-1,
            legend cell align={left},
        ]

            \addplot[only marks,color=cd_plot,mark=diamond*,error bars/.cd,x dir=both,x explicit] coordinates {
                (0.349925136,5) +- (0.15354648651,0) 
                (1.723251472,4) +- (0.94678665034,0) 
                (0.184416428,3) +- (0.09429510047,0) 
                (0.285179484,2) +- (0.10223667976,0) 
                (0.106880366,1) +- (0.06185044114,0) 
                (0,0) +- (0,0) 
            };

            \addplot[only marks,color=mae_plot,mark=square*,error bars/.cd,x dir=both,x explicit] coordinates {
                (1.432540355,5) +- (0.49980258704,0) 
                (2.175665678,4) +- (0.52025792223,0) 
                (0.543719084,3) +- (0.18804246770,0) 
                (0.978669762,2) +- (0.31115563325,0) 
                (0.440802884,1) +- (0.16275757997,0) 
                (0,0) +- (0,0) 
            };

            \addplot[only marks,color=mse_plot,mark=triangle*,error bars/.cd,x dir=both,x explicit] coordinates {
                (29.10329969/10,5) +- (16.3817447216/10,0) 
                (30.11770062/10,4) +- (14.2329987137/10,0) 
                (4.638681825/10,3) +- (3.05137111863/10,0) 
                (16.10290359/10,2) +- (9.31805084431/10,0) 
                (5.098773228/10,1) +- (3.32631586608/10,0) 
                (0,0) +- (0,0) 
            };

            \legend{CD\!~\![m],MAE\!~\![m],MSE\!~\!$[10m^2]$}
        \end{axis}

        \begin{axis}[
            name=plot2,at={($(plot1.east)+(0.4cm,0)$)},
            anchor=west,
            height=4cm,
            width=5.2cm,
            ytick=data,
            yticklabels=\empty,
            xlabel=Metric Scores,
            xmin=0,
            xmax=1,
            grid=both,
            legend style={at={(0.5,1.02)},anchor=south,draw=none,nodes={scale=0.8, transform shape}},
            legend columns=-1,
            legend cell align={left},
        ]

            \addplot[misc_plot,mark=square*,only marks,error bars/.cd,x dir=both,x explicit] coordinates {
                (0.001495700403,5) +- (0.001264764300,0) 
                (0.257766407048,4) +- (0.025629698682,0) 
                (0.205078136596,3) +- (0.025664409263,0) 
                (0.028375717138,2) +- (0.012577294951,0) 
                (0.021826540240,1) +- (0.010947225242,0) 
                (0.015455430647,0) +- (0.008232826823,0) 
            };

            \addplot[real_plot,mark=*,only marks,error bars/.cd,x dir=both,x explicit] coordinates {
                (0.601425613716,5) +- (0.049490766262,0) 
                (0.641733616782,4) +- (0.033728459606,0) 
                (0.689199649825,3) +- (0.028951115422,0) 
                (0.807546045674,2) +- (0.023138403416,0) 
                (0.872251151886,1) +- (0.021551970891,0) 
                (0.835584440616,0) +- (0.025164888800,0) 
            };

            \addplot[syn_plot,mark=triangle*,only marks,error bars/.cd,x dir=both,x explicit] coordinates {
                (0.397078688088,5) +- (0.049594748482,0) 
                (0.100499975680,4) +- (0.016313186974,0) 
                (0.105722212426,3) +- (0.014408363828,0) 
                (0.164078238482,2) +- (0.021816946973,0) 
                (0.105922308697,1) +- (0.018720700562,0) 
                (0.148960127605,0) +- (0.023068917684,0) 
            };

            \legend{$S^\Misc{}$,$S^\Real{}$,$S^\Syn{}$}

        \end{axis}

        \begin{axis}[
            name=plot3,at={($(plot2.east)+(0.4cm,0)$)},
            anchor=west,
            height=4cm,
            width=5.2cm,
            ytick=data,
            yticklabels=\empty,
            xlabel=Segmentation mIoU,
            xmin=0.35,
            xmax=0.6,
            grid=both,
            legend style={at={(0.5,1.02)},anchor=south,draw=none,nodes={scale=0.8, transform shape}},
            legend columns=2,
            legend cell align={left},
        ]

            \addplot[color=mae_plot,mark=diamond*,only marks,mark options={solid,fill opacity=0.5},error bars/.cd,x dir=both,x explicit]
            coordinates {
                (0.4149,5) +- (0.0286,0) 
                (0.4721,4) +- (0.0059,0) 
                (0.4872,3) +- (0.0505,0) 
                (0.4881,2) +- (0.0438,0) 
                (0.4907,1) +- (0.0550,0) 
                (0.5142,0) +- (0.0265,0) 
            };

            \addplot[color=mse_plot,mark=*,only marks,mark options={solid,fill opacity=0.5},error bars/.cd,x dir=both,x explicit]
            coordinates {
                (0.4926,5) +- (0.0017,0) 
                (0.4663,4) +- (0.0013,0) 
                (0.5221,3) +- (0.0013,0) 
                (0.5192,2) +- (0.0010,0) 
                (0.5275,1) +- (0.0007,0) 
                (0.5259,0) +- (0.0013,0) 
            };

            \legend{DarkNet21,SqueezeSegV2}

        \end{axis}

        \par} 

    \end{tikzpicture}

    \caption{
        \textbf{Metric Scores for Up-Sampling Methods}:
        The vertical axis lists five methods to perform $4\times$ LiDAR scan up-sampling and the high-resolution target data~(``KITTI'').
        The left plot shows the reconstructions errors of different baseline measures.
        The middle plot shows the three parts of our realism measure.
        The right plot shows the semantic segmentation results on the original KITTI dataset of a segmentation model trained with the data generated from the respective row.
        The methods are ordered from top to bottom by increasing human visual judgment ratings.
    }
    \label{fig:experiments_upsampling_metric_4x}
\end{figure*}
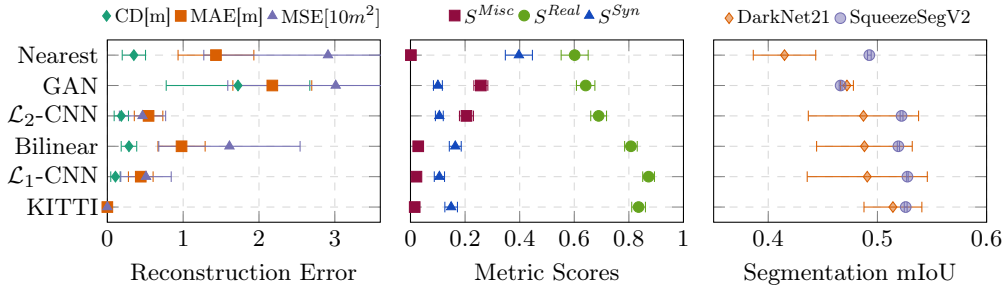

\section{Metric Application}
\label{sec:application}

In this section we demonstrate how our realism measure ranks different datasets generated by neural networks.
We then compare these results to our baseline evaluation measures (introduced in~\secref{sec:setup_baseline}), and analyse the resulting performance of a segmentation network.

\figref{fig:experiments_upsampling_metric_4x}~is divided into three parts horizontally.
The leftmost plot shows the baseline metrics, the middle shows the results of our metric network, the rightmost plot shows the segmentation performance.
The vertical axis on the left lists five different versions of KITTI data, generated as explained in~\secref{sec:setup_upsampling}.
For the displayed segmentation results, different versions of the same model were trained with each of the datasets and then evaluated on the original KITTI data.

The realism score for the original KITTI is displayed for reference and has reconstruction errors of zero.
The methods are ranked from top to bottom by increasing realism as approximately perceived by humans\footnote{It is not clear how to rank the \textit{nearest neighbor interpolation} here, since its appearance is completely different to the others. Therefore we simply placed it according to its metric score.}.
In general, the baseline metrics show a tendency but no clear correlation to the degree of realism and struggle to produce an unambiguous ordering of the methods.
Our realism score, on the other hand, sorts the up-sampling methods according to human visual judgment.
These results align with the ones in~\citep{Triess2019IV}, which shows that a low reconstruction error does not necessarily imply high realism in the generated outputs.
This is the main reason for the emergence of perceptual losses in recent years~\cite{JohnsonECCV2016,LedigCVPR2017}.

\begin{table*}
    \centering

    \caption{
        \textbf{Semantic Segmentation Performance}:
        The table lists the evaluation results of the DarkNet21 model for point-wise semantic segmentation.
        For each row, the model is trained on the respective dataset which corresponds to a high-resolution KITTI variation generated from low-resolution data.
        The evaluation results are all reported on the validation split of the original KITTI data.
        All numbers in the table are given in \%.
        Best results are shown in \textbf{bold}, second best in \textit{italic}.
    }
    \label{tab:experiments_semseg}

    {\small
    \begin{tabularx}{\linewidth}{X | cc | ccccccccc}
        \toprule 
        KITTI Version &
        \rotatebox{90}{accuracy} &
        \rotatebox{90}{mean IoU} &
        \rotatebox{90}{\colorcube[sem_person] person} &
        \rotatebox{90}{\colorcube[sem_twowheeler] two-wheeler} &
        \rotatebox{90}{\colorcube[sem_largevehicle] large-vehicle} &
        \rotatebox{90}{\colorcube[sem_vehicle] vehicle} &
        \rotatebox{90}{\colorcube[sem_road] road} &
        \rotatebox{90}{\colorcube[sem_sidewalk] sidewalk} &
        \rotatebox{90}{\colorcube[sem_terrain] terrain} &
        \rotatebox{90}{\colorcube[sem_construction] construction} &
        \rotatebox{90}{\colorcube[sem_vegetation] vegetation} \\
        \midrule 
        \multicolumn{3}{l}{Class frequency $\frac{\text{train}}{\text{val}}$} &
        $\frac{0.05}{0.16}$ & $\frac{0.06}{0.12}$ & $\frac{0.22}{0.10}$ & $\frac{4.4}{6.2}$ & $\frac{21.7}{18.9}$ & $\frac{14.5}{12.1}$ & $\frac{07.7}{12.9}$ & $\frac{21.3}{14.3}$ & $\frac{26.8}{30.3}$ \\
        \midrule 

        Nearest & 76.0 & 41.5 &
        13.7 & 3.3 & \textit{0.5} & \textit{89.0} & 78.6 & 47.6 & 24.5 & 56.9 & 71.8 \\

        GAN & 81.1 & 47.2 &
        \textit{16.5} & \textbf{9.3} & 0.2 & 88.0 & \textbf{86.5} & \textbf{71.1} & 62.5 & 69.1 & 79.5 \\

        L2-CNN & 82.8 & 48.7 &
        13.5 & 1.9 & \textbf{1.3} & \textbf{90.0} & 82.7 & 62.0 & 66.1 & 69.8 & 79.1 \\

        Bilinear & 83.2 & 48.8 &
        12.5 & 4.5 & 0.3 & 88.6 & 84.5 & 67.4 & \textit{69.0} & \textit{72.0} & \textit{81.1} \\

        L1-CNN & \textit{83.5} & \textit{49.1} &
        10.9 & 2.1 & 0.2 & 86.2 & \textit{84.7} & \textit{68.1} & 65.4 & 71.3 & 79.5 \\

        \midrule 
        Original & \textbf{84.9} & \textbf{51.4} &
        \textbf{20.7} & \textit{6.5} & \textit{0.5} & 87.0 & 84.5 & 67.7 & \textbf{69.2} & \textbf{73.7} & \textbf{82.6} \\
        \bottomrule 
    \end{tabularx}
    \par}

\end{table*}

\begin{figure*}
    \centering

    \begin{tabularx}{\linewidth}{@{\hskip 1pt}c@{\hskip 1pt}c@{\hskip 1pt}c@{\hskip 1pt}c@{\hskip 1pt}c@{\hskip 1pt}c@{\hskip 1pt}c@{\hskip 1pt}}
        GT Labels & KITTI & $\mathcal{L}_1$-CNN & Bilinear & $\mathcal{L}_2$-CNN & GAN & Nearest \\

        &
        \frame{\includegraphics[width=0.14\linewidth,angle=90,origin=c]{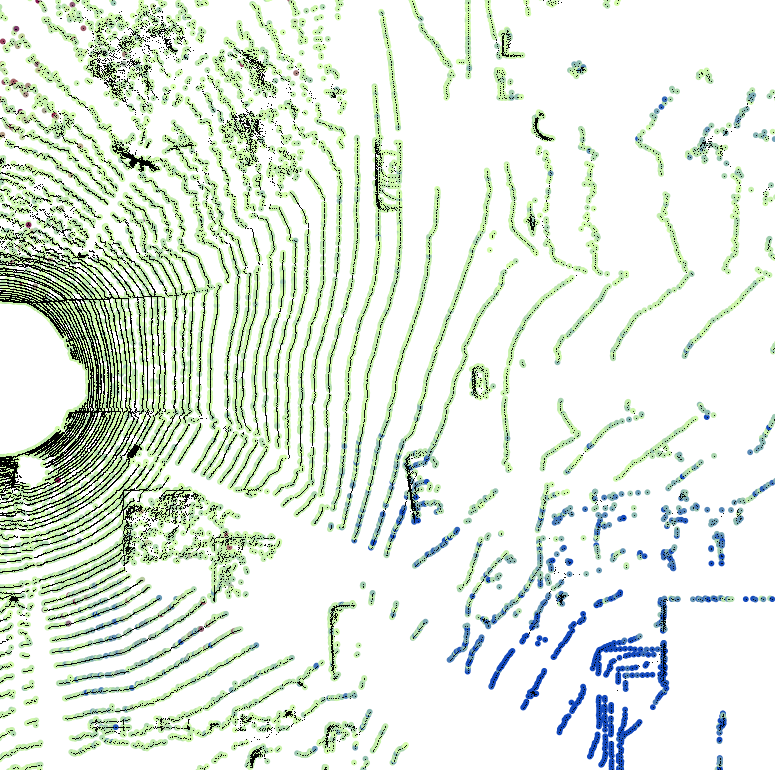}} &
        \frame{\includegraphics[width=0.14\linewidth,angle=90,origin=c]{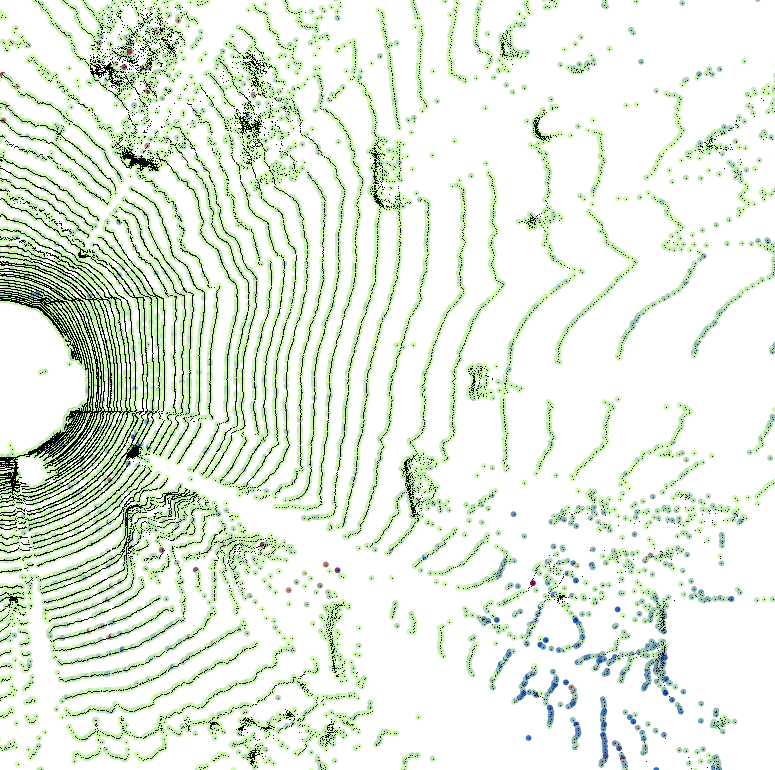}} &
        \frame{\includegraphics[width=0.14\linewidth,angle=90,origin=c]{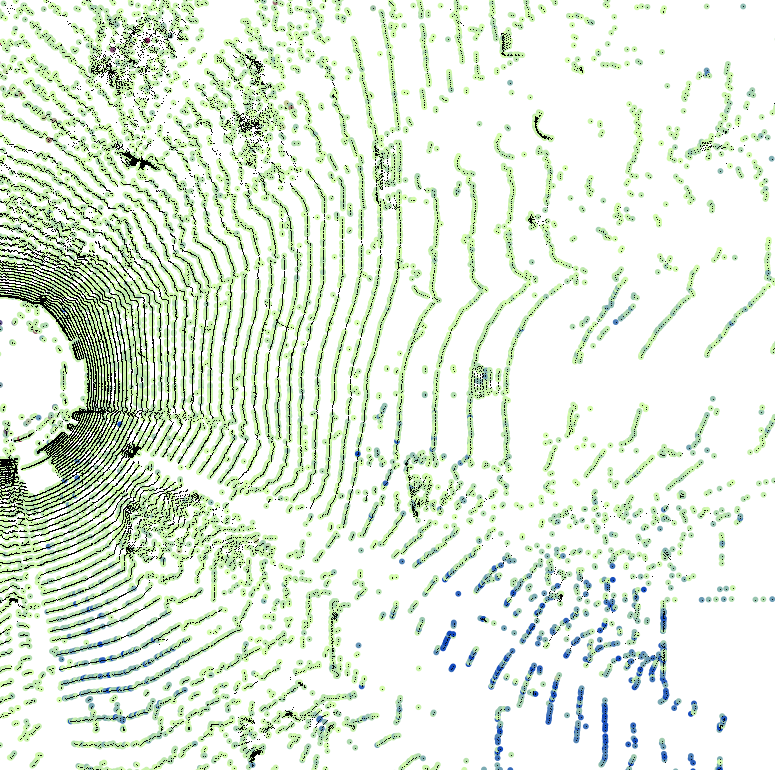}} &
        \frame{\includegraphics[width=0.14\linewidth,angle=90,origin=c]{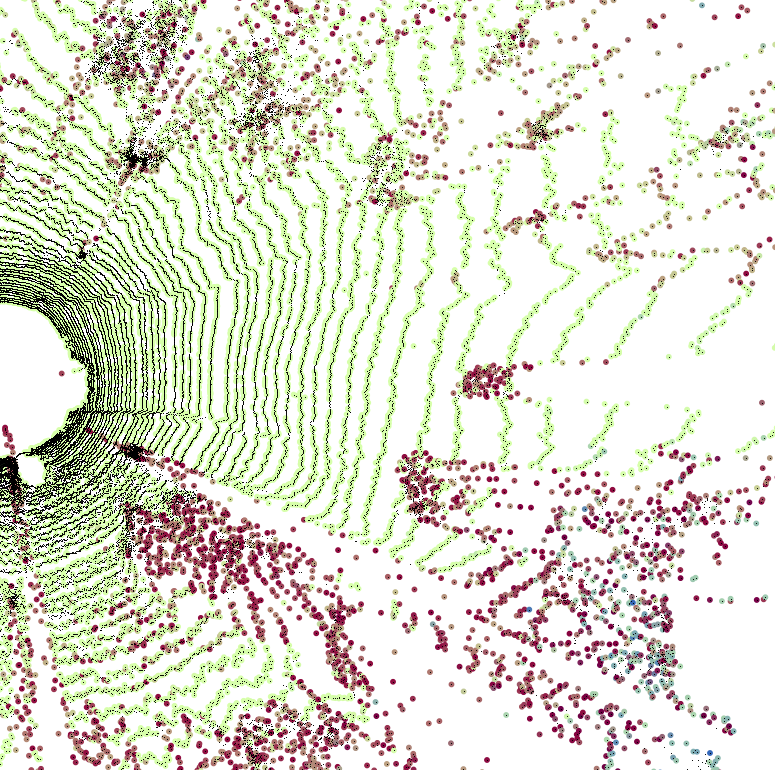}} &
        \frame{\includegraphics[width=0.14\linewidth,angle=90,origin=c]{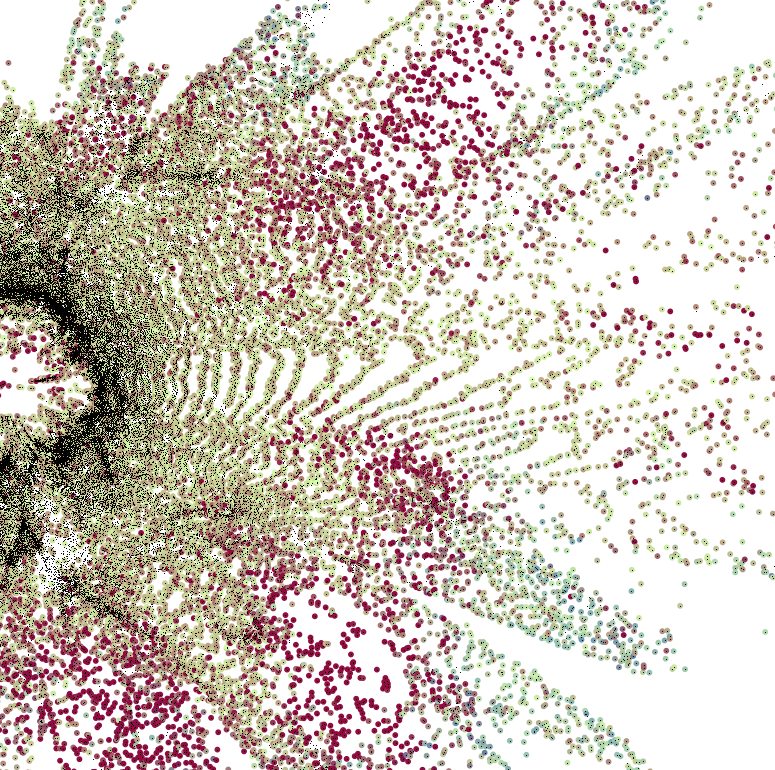}} &
        \frame{\includegraphics[width=0.14\linewidth,angle=90,origin=c]{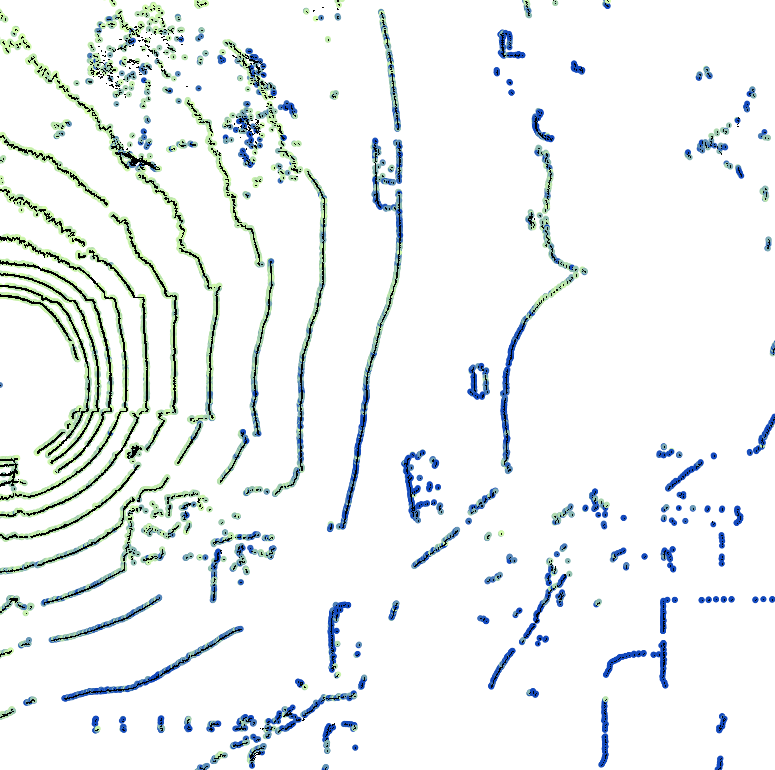}} \\

        &
        \frame{\includegraphics[width=0.14\linewidth]{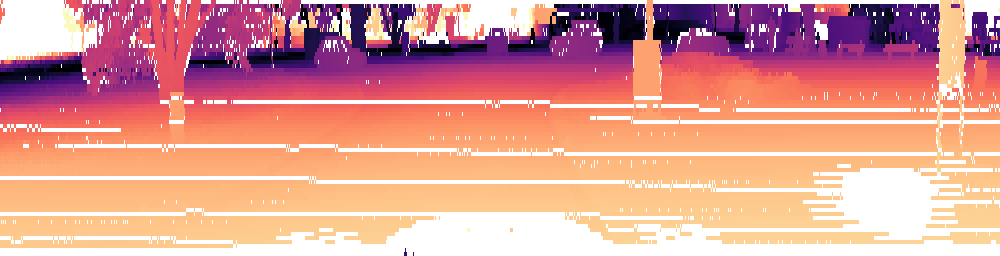}} &
        \frame{\includegraphics[width=0.14\linewidth]{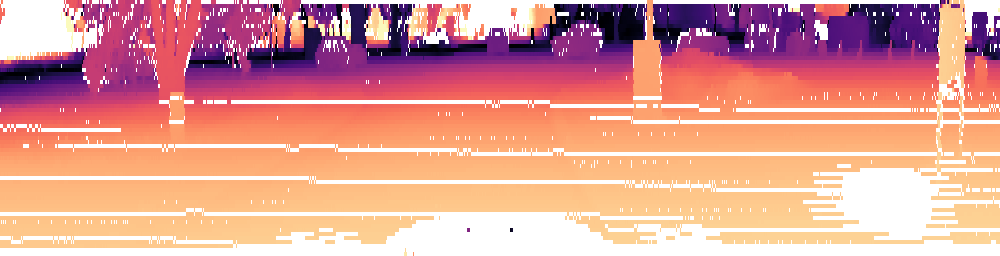}} &
        \frame{\includegraphics[width=0.14\linewidth]{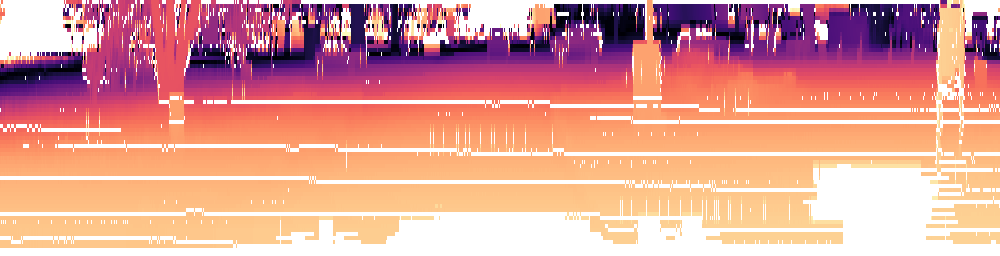}} &
        \frame{\includegraphics[width=0.14\linewidth]{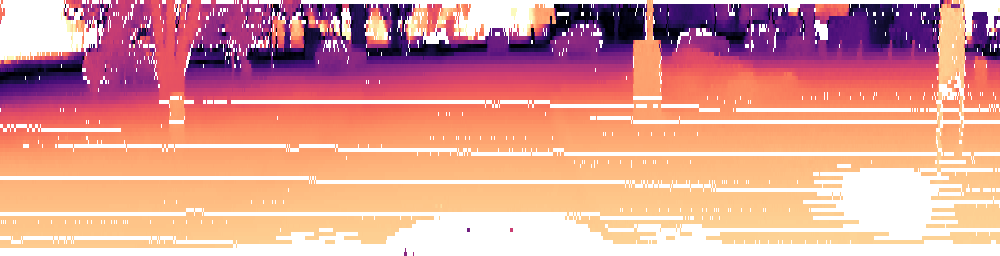}} &
        \frame{\includegraphics[width=0.14\linewidth]{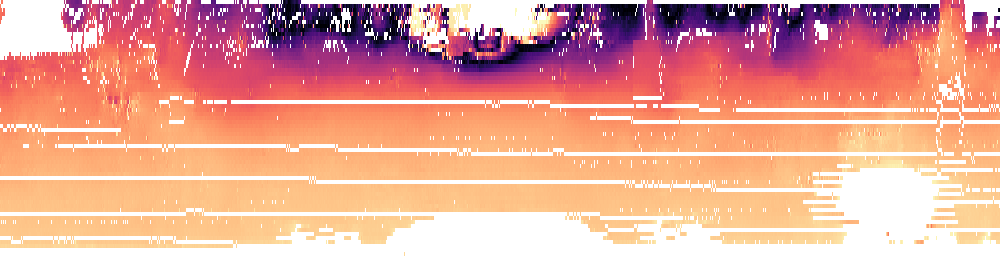}} &
        \frame{\includegraphics[width=0.14\linewidth]{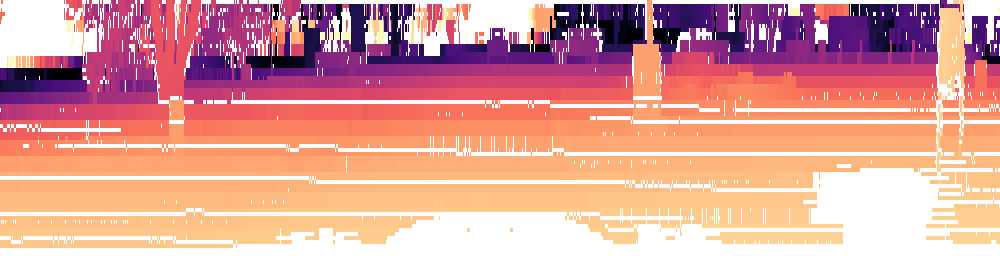}} \\

        &
        &
        \frame{\includegraphics[width=0.14\linewidth]{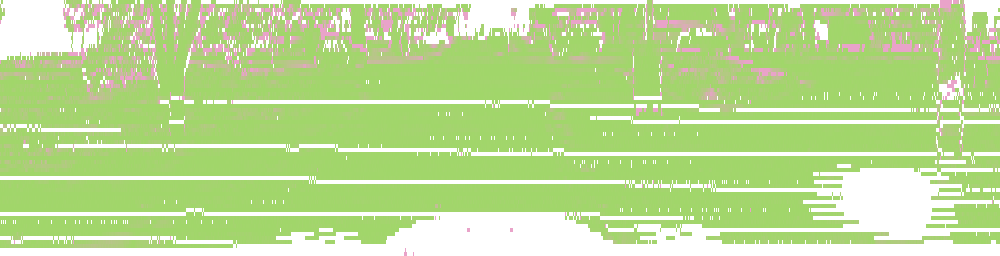}} &
        \frame{\includegraphics[width=0.14\linewidth]{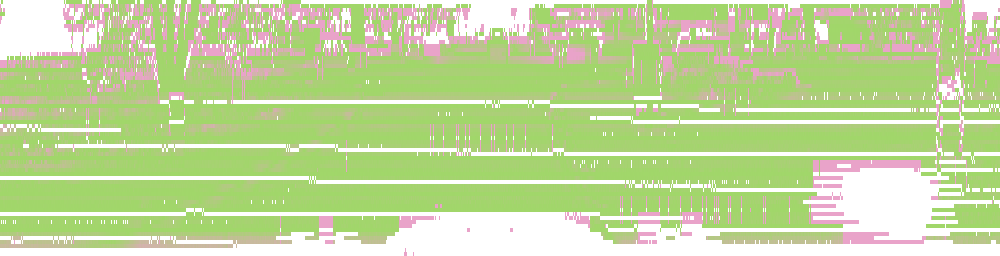}} &
        \frame{\includegraphics[width=0.14\linewidth]{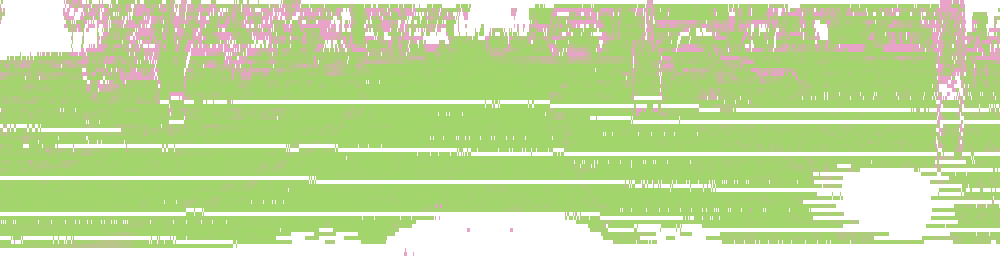}} &
        \frame{\includegraphics[width=0.14\linewidth]{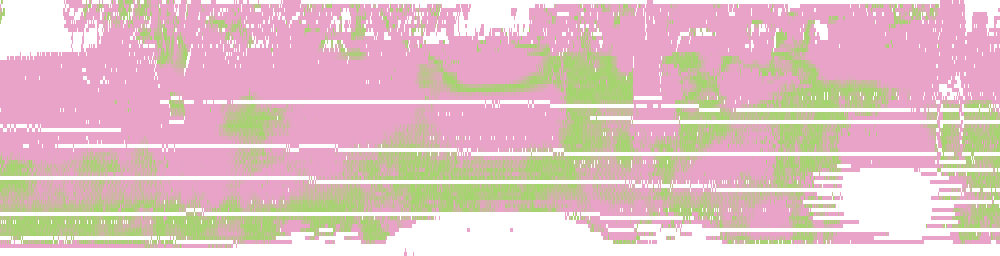}} &
        \frame{\includegraphics[width=0.14\linewidth]{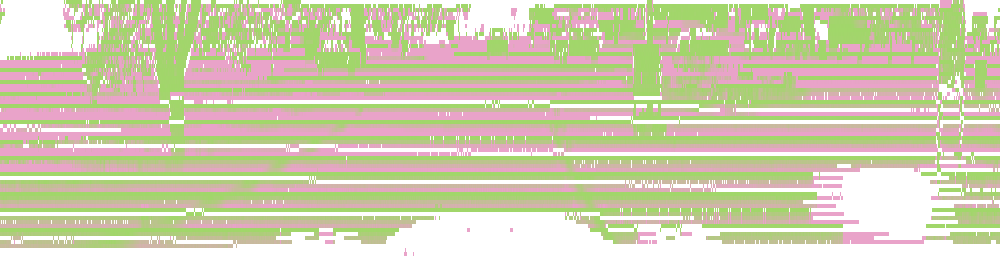}} \\

        \frame{\includegraphics[width=0.14\linewidth]{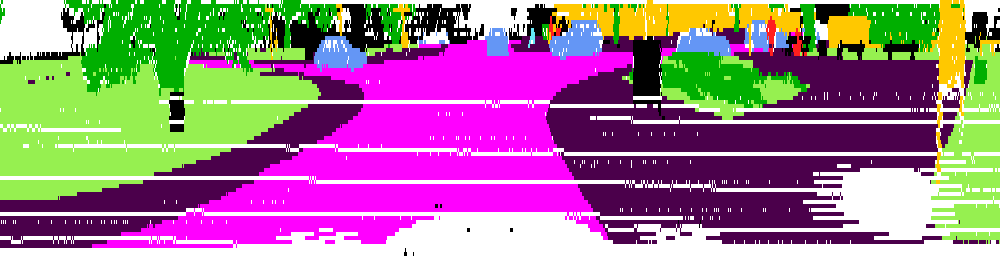}} &
        \frame{\includegraphics[width=0.14\linewidth]{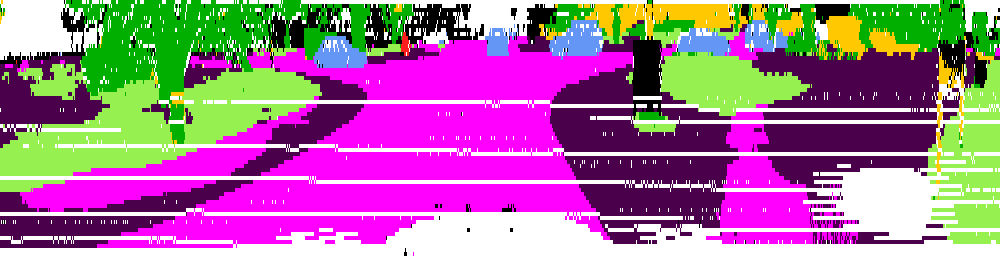}} &
        \frame{\includegraphics[width=0.14\linewidth]{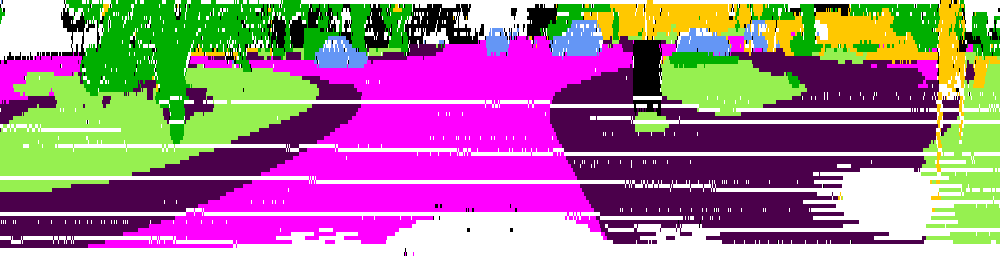}} &
        \frame{\includegraphics[width=0.14\linewidth]{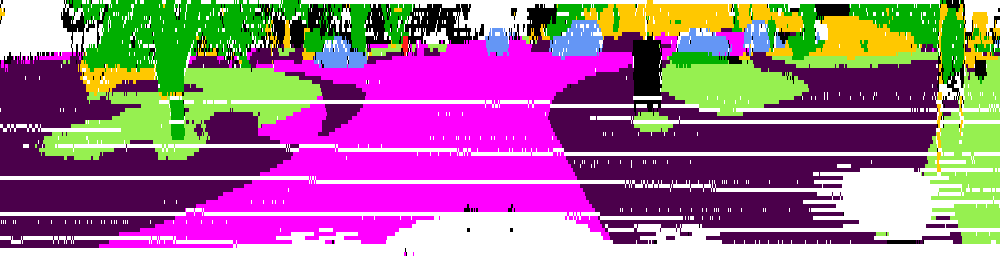}} &
        \frame{\includegraphics[width=0.14\linewidth]{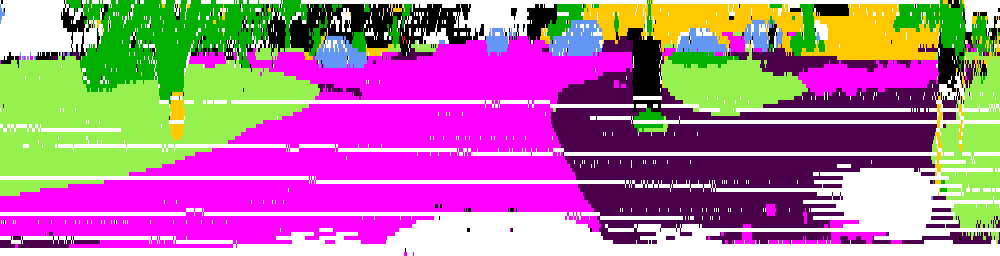}} &
        \frame{\includegraphics[width=0.14\linewidth]{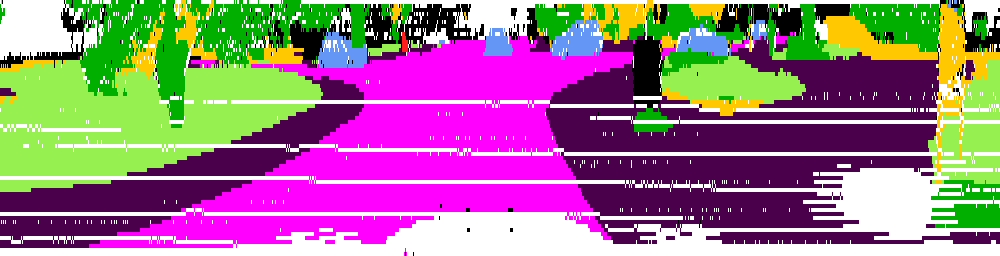}} &
        \frame{\includegraphics[width=0.14\linewidth]{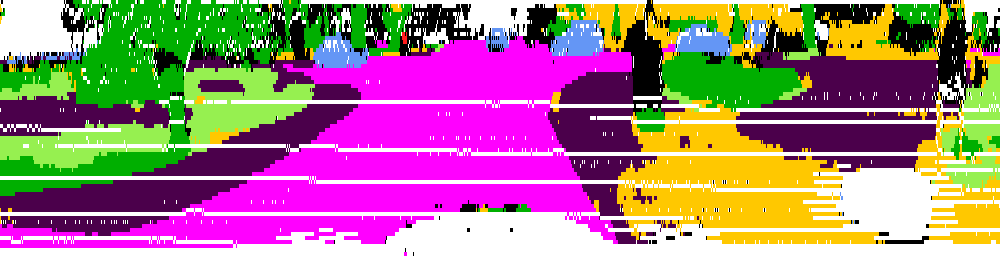}} \\

        &
        \frame{\includegraphics[width=0.14\linewidth]{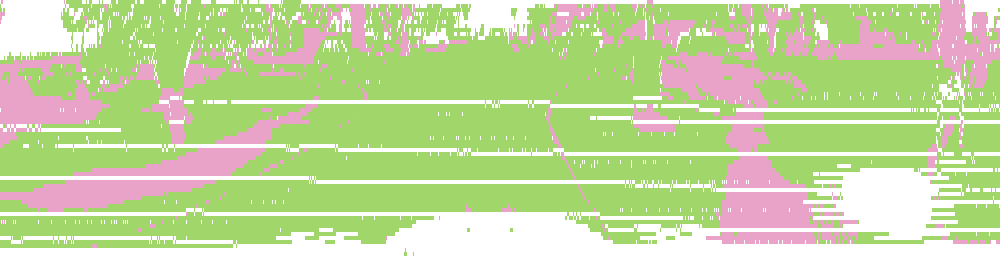}} &
        \frame{\includegraphics[width=0.14\linewidth]{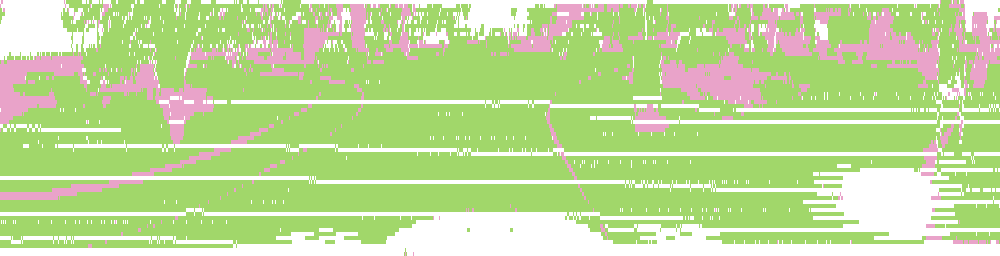}} &
        \frame{\includegraphics[width=0.14\linewidth]{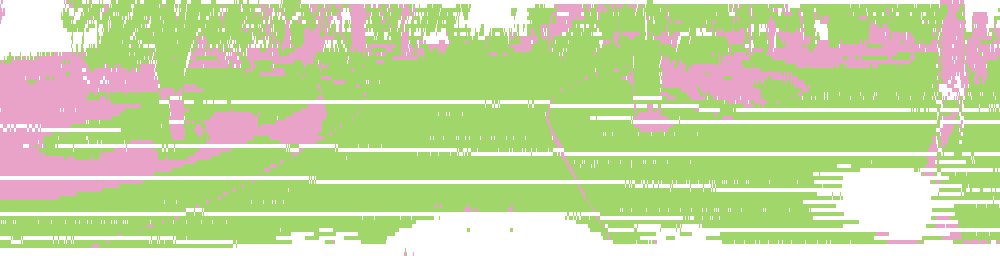}} &
        \frame{\includegraphics[width=0.14\linewidth]{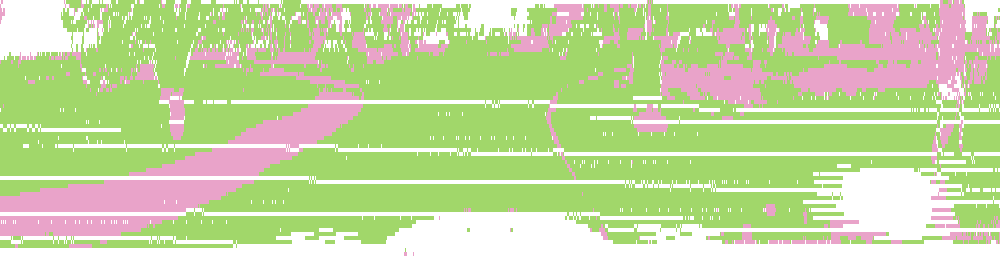}} &
        \frame{\includegraphics[width=0.14\linewidth]{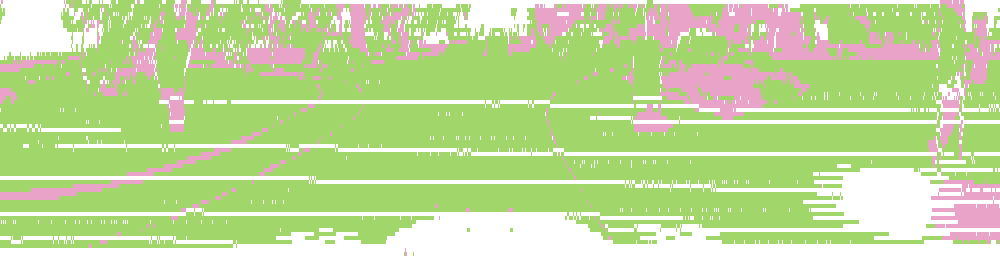}} &
        \frame{\includegraphics[width=0.14\linewidth]{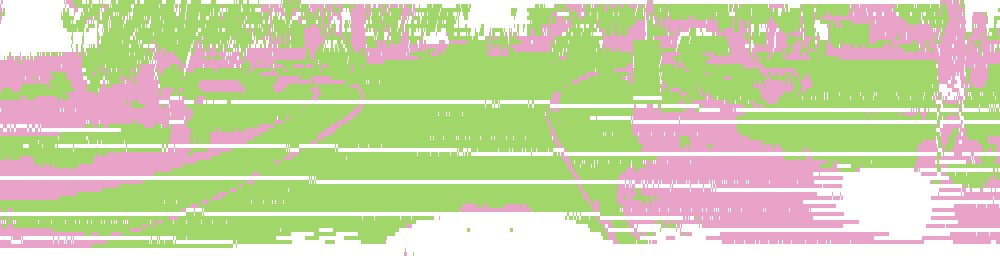}} \\

        \frame{\includegraphics[width=0.14\linewidth,angle=90,origin=c]{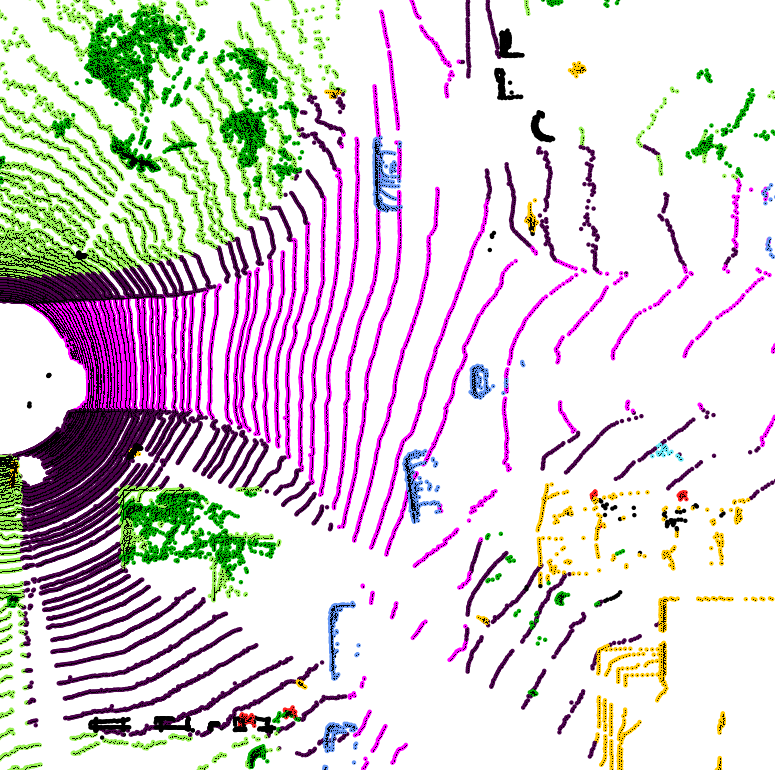}} &
        \frame{\includegraphics[width=0.14\linewidth,angle=90,origin=c]{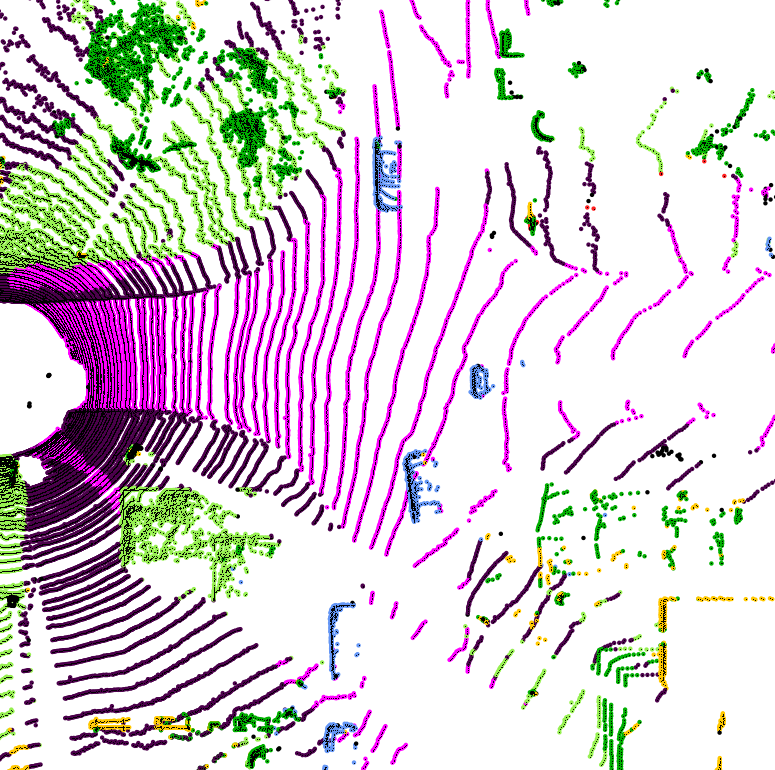}} &
        \frame{\includegraphics[width=0.14\linewidth,angle=90,origin=c]{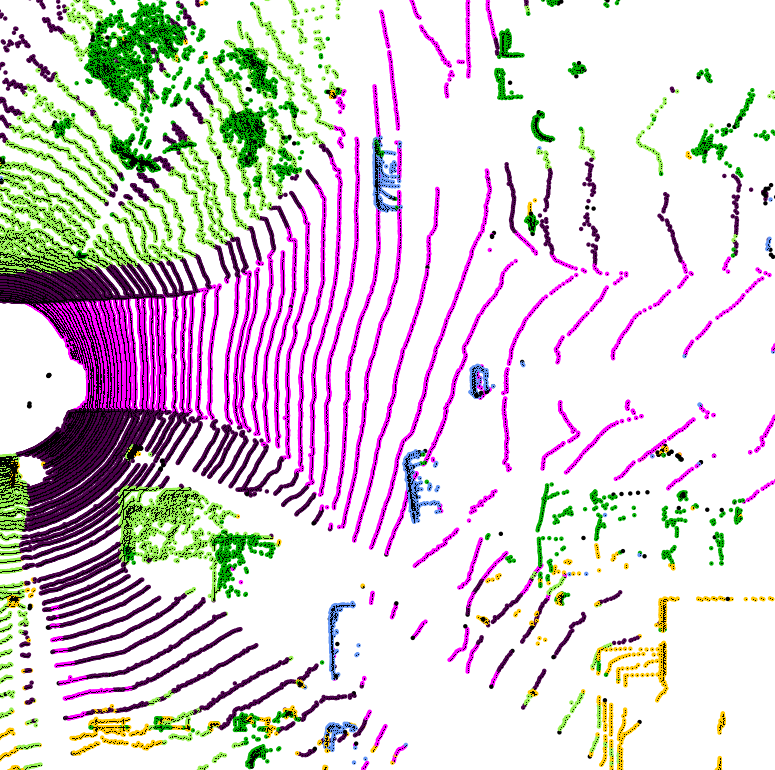}} &
        \frame{\includegraphics[width=0.14\linewidth,angle=90,origin=c]{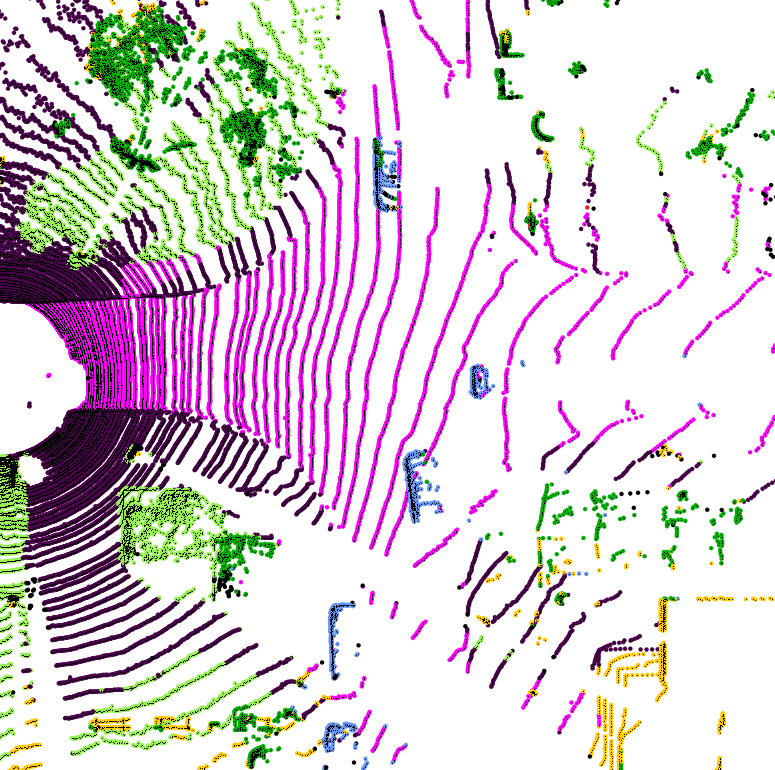}} &
        \frame{\includegraphics[width=0.14\linewidth,angle=90,origin=c]{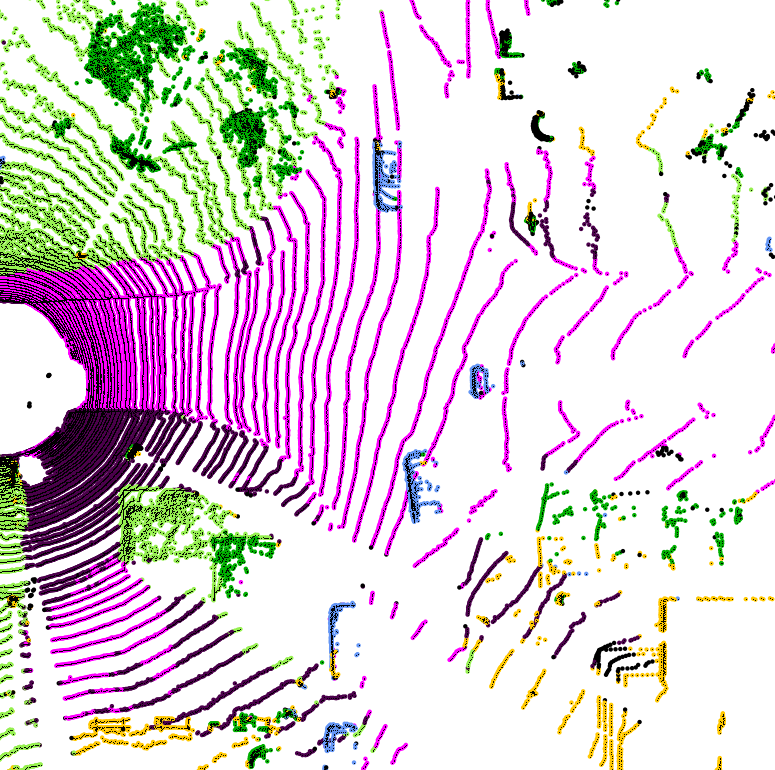}} &
        \frame{\includegraphics[width=0.14\linewidth,angle=90,origin=c]{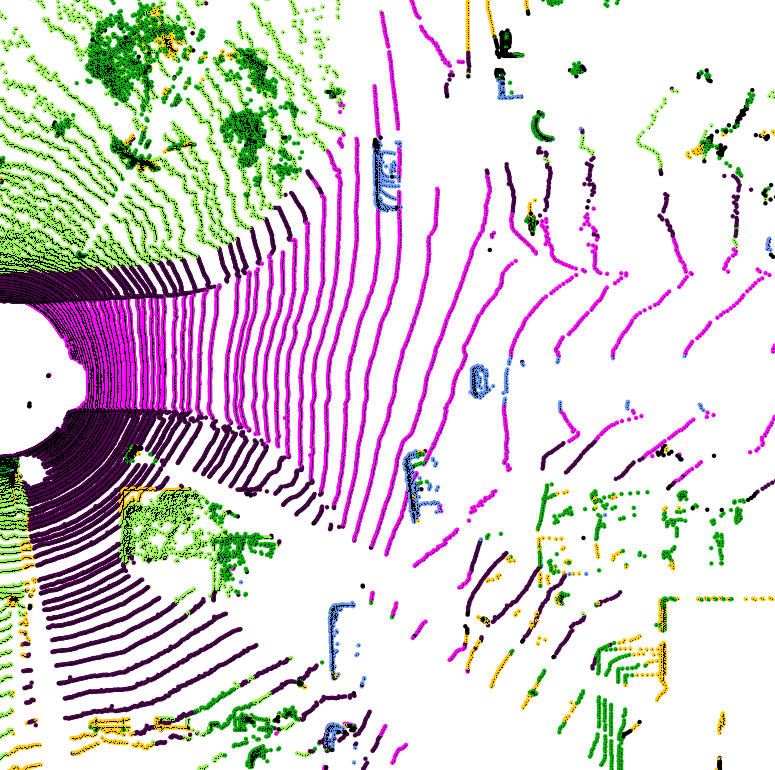}} &
        \frame{\includegraphics[width=0.14\linewidth,angle=90,origin=c]{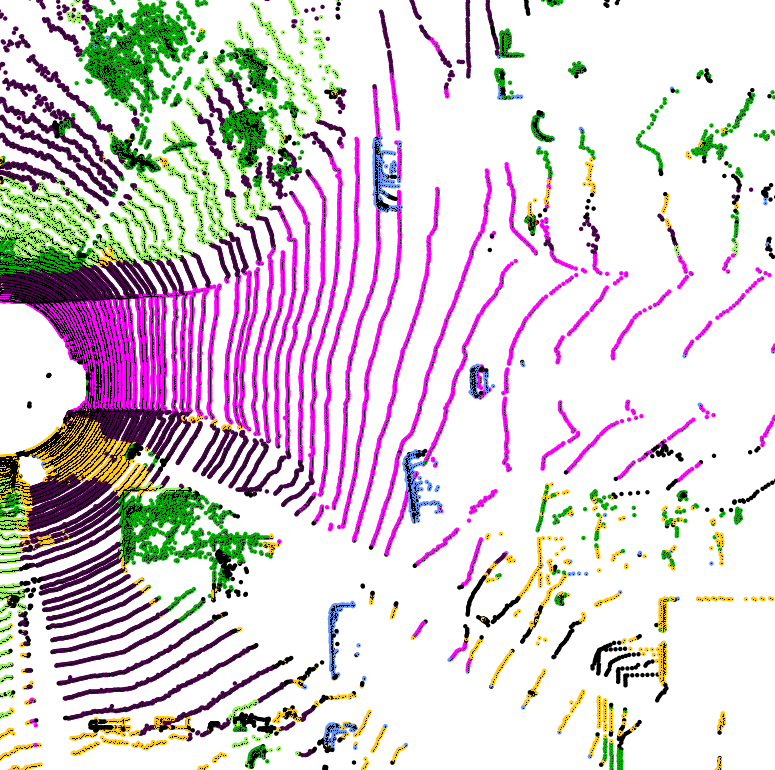}} \\
    \end{tabularx}

    \caption{
        \textbf{Qualitative Up-Sampling and Segmentation Results}:
        The first row shows the metric results on an up-sampled KITTI scene.
        The original scan is shown in column ``KITTI''.
        The colors are soft interpolations of \Real{}~\colorcube[real_img], \Syn{}~\colorcube[syn_img], and \Misc{}~\colorcube[misc_img].
        The second row shows the color-coded depth projection of the point cloud.
        The third row shows the relative error between the generated sample and the original high resolution sample from column ``KITTI''.
        A pixel is green~\colorcube[plot_low] if the error is 0\% and pink~\colorcube[plot_high] if the error is higher than 10\%, all values in between are linearly interpolated.
        The fourth and sixth row show the segmentation results of a model trained on the respective up-sampled data.
        A legend of the semantic colors is provided in~\tabref{tab:experiments_semseg}.
        The ground truth semantic labels are shown in the leftmost column.
        For better comparison, the fifth column shows correctly classified pixels in green~\colorcube[plot_low] and wrong classifications in pink~\colorcube[plot_high].
        The visualized sample is from the validation split and was neither used to train the metric, nor the segmentation network.
    }
    \label{fig:experiments_qualitative}
\end{figure*}

The upper row of \figref{fig:experiments_qualitative} shows an example scene for all up-sampling versions with their obtained scores.
The $\mathcal{L}_1$-CNN produces an almost perfect version of the original high-resolution data, only with some noise at object boundaries.
Bilinear interpolation works very well on large surfaces, but produced single noise points especially in regions where the LiDAR usually receives no return, \eg windows.
The $\mathcal{L}_2$-CNN can reconstruct the outlines of the scene, but suffers from high noise throughout the entire point cloud.
Similarly, the up-sampling GAN suffers from high noise, but often is not able to reconstruct the outlines of the scene and forms random point clusters instead of clear objects.
The nearest neighbor interpolation causes vertically stretched objects, which works fine for walls, poles, and other vertical objects, but fails for the ground.

These differences also cause different behavior in downstream perception in the target domain when the generated data is used for training.
The rightmost plot in \figref{fig:experiments_upsampling_metric_4x} shows the overall results, while \tabref{tab:experiments_semseg} shows class-wise results.
Additionally, the bottom row of \figref{fig:experiments_qualitative} visualizes segmented example point clouds produced by the models trained with the respective data.
Both segmentation models show similar trends for the order of the up-sampling methods as the realism metric.
The slightly higher \Real{} score for $\mathcal{L}_1$-CNN than for the original KITTI data can also be seen in the segmentation score of the SqueezeSegV2 model, but is neither significant nor does it behave in the same way for DarkNet21.
Also the SqueezeSegV2 behavior on the nearest neighbor up-sampling is not equal to those of DarkNet21 and the metric.
It can be assumed that two effects lead to this different behavior:
First, as mentioned in footnote~1, it is not clear how exactly the nearest neighbor interpolation should be judged in terms of realism.
Second, SqueezeSegV2 exhibits almost no variance on its performance scores.
The combination of these two effects could cause the difference in behavior, but it is not quite clear how and therefore needs further investigation which is left for future work.

The class-wise results in \tabref{tab:experiments_semseg} show that $\mathcal{L}_2$-CNN and GAN achieve quite good results for dynamic objects.
At the same time, it is very hard to tell which of the point clusters in the 3D visualization of~\figref{fig:experiments_qualitative} belong to these objects.
This raises the question why training with this highly distorted data achieves such good performance in the target domain.
The question can be answered by looking at the projected LiDAR scan.
Here it becomes visible that even regions that suffer from high noise can still be approximately detected by their edge outlines in the projection.
The third row of~\figref{fig:experiments_qualitative} shows the point-wise relative error between the generated and the target point cloud with the error being clipped to a maximum of~10\%.
Even for the appearing noisy $\mathcal{L}_2$-CNN version, relative errors are quite low and therefore outlines are clearly visible in the depth projection (second row).
We find that this is an indication that the segmentation model is not influenced by local noise perturbations, but rather learns a more generalized appearance of the object shapes.

\section{Discussion}
\label{sec:discussion}

Our experiments show a correlation between measured training data realism and final perception performance.
Qualitatively however, the segmentation performance seems to be less affected by reduced point cloud realism than expected by judging from the 3D images.
We believe that this is caused by the selected architectures of the up-sampling and segmentation models.
The segmentation networks operate on the 2D projections of the point clouds which is similar to the projection space used for up-sampling.
Even though objects are blurred and unrecognizable when the GAN up-sampling is displayed as raw 3D data, objects shapes are still detectable on the 2D projections.
We make two considerations from this observation:

First, visual judgment is highly dependent on the chosen data representation and their visualization.
This is an important reason to use such a quantitative metric as ours on a large amount of data.
Second, we believe that our metric might be more reliable to estimate the performance of downstream tasks operating on 3D space.

A major concept to keep in mind is the difference between domain gap and realism.
If the task to solve is to train a method for KITTI-to-nuScenes adaptation, then both the target and the source domain are \Real{}.
Our metric can be used to rule out any unrealistic data compositions that form in the transition between those two datasets, \eg{} while training a domain adaptation method.
However, if the method is just outputting an identity function, the realism would be at maximum, while the domain gap still causes bad perception performance in the target domain.
Therefore, tasks are only in parts dependent on the realism of the data and domain gaps have to be measured differently.

\section{Conclusion}
\label{sec:conclusion}

This paper presented a novel metric to quantify the degree of realism of local regions in LiDAR point clouds.
Through adversarial learning, we obtain a feature encoding that is able to adequately capture data realism more generally instead of focusing on dataset-specific characteristic.
In extensive experiments, we demonstrated the reliability and applicability of our metric on unseen data.
The predictions of our method correlate well with visual judgment, unlike reconstruction errors serving only as a proxy for realism.
In addition, we investigated the influence of data realism on a downstream perception task.

Future work includes to design a generative model that uses synthetic data, \eg{} CARLA, as input to generate realistic real-world data, \eg{} KITTI.
Our metric is used in this setup to find the optimal point where the generated data actually improves the downstream perception performance of a segmentation or object detection model.

\backmatter 







\bibliography{bibliography/journals_long,bibliography/references}

\FloatBarrier
\begin{appendices} 

\section{Metric Implementation Details}
\label{app:implementation}


\tabref{tab:implementation_metric}~lists all layers, inputs, and operations of our \ac{dnn} architecture.
We use TensorFlow to implement online data processing, neural network weight optimization, and network inference.
The implementation is oriented on the original PointNet++ implementation~\citep{Qi2017NIPS}\footnote{PointNet++ code \url{https://github.com/charlesq34/pointnet2}}.
The Adam optimizer is used for optimization.
We use an initial learning rate of $1e^{-3}$ with exponential warm-up and decay.

The classifier outputs the scores for each of the $U_C=3$ categories, namely \Real{}, \Syn{}, \Misc{}.
The adversary for \Real{} has $U_A^\Real{}=2$ output channels, for KITTI and nuScenes.
The \Syn{} adversary outputs $U_A^\Syn{}=2$ scores for CARLA and GeoSet.
For the \Misc{} category, the respective adversary has $U_A^\Misc{}=3$ outputs, for Misc~{1,2,3}.
Implementation-wise, all adversaries have the full seven output channels for all datasets.
The category split is implemented as a class weighting when computing the loss from the adversary output, such that the loss becomes zero if the input does not origin from within the respective category.
We found this the easiest and most stable way to implement the desired behavior in TensorFlow graph mode.

\begin{table*}
    \centering

    \caption{
        \textbf{Network Architecture}:
        Detailed network architecture and input format definition.
        The ID of each row is used to reference the output of the row. $\uparrow$ indicates that the layer directly above is an input.
        $N$ denotes the number of LiDAR measurements.
        $Q_j$ are the number of query points at abstraction level $j$.
        $K_j$ are the number of nearest neighbors to search at abstraction level $j$.
        $U$ are the number of output units of the classifier and the adversaries.
    }
    \label{tab:implementation_metric}

    \begin{tabularx}{\linewidth}{rlll X}
        \toprule
        \textbf{ID} & \textbf{Inputs} & \textbf{Operation} & \textbf{Output Shape} & \textbf{Description} \\
        \midrule
        1 & LiDAR & $x$, $y$, $z$ & $[N \times 3]$ & Position of each point relative to sensor origin \\
        \midrule
        \multicolumn{5}{c}{\textbf{Feature Extractor: Abstraction Module 1}} \\
        \midrule
        2 & $\uparrow$, $Q_1$ & Farthest point sampling & $[2048]$ & Indices of $Q_1$ query points \\
        3 & 1, $\uparrow$ & Group & $[2048 \times 3]$ & Grouped sampled points \\
        4 & 1, 2, $K_1$ & Nearest neighbor search & $[2048 \times 20]$ & Indices of the $K_1$ nearest neigbors per query \\
        5 & 1, 2, $\uparrow$ & Group & $[2048 \times 20 \times 3]$ & Grouped neighborhoods \\
        6 & $\uparrow$ & Neighborhood normalization & $[2048 \times 20 \times 3]$ & Translation normalization towards query point \\
        7 & $\uparrow$ & (Conv+LeakyReLU) $\times\!2$ & $[2048 \times 20 \times 64]$ & Kernel size $1\!\times\!1$, stride 1 \\
        8 & $\uparrow$ & Conv+LeakyReLU & $[2048 \times 20 \times 128]$ & Kernel size $1\!\times\!1$, stride 1 \\
        9 & $\uparrow$ & ReduceMax & $[2048 \times 128]$ & Maximum over neighborhood features \\
        \midrule
        \multicolumn{5}{c}{\textbf{Feature Extractor: Abstraction Module 2}} \\
        \midrule
        10 & 3, $Q_2$ & Farthest point sampling & $[256]$ & Indices of $Q_2$ query points \\
        11 & 3, 10, $K_2$ & Nearest neighbor search & $[256 \times 10]$ & Indices of the $K_2$ nearest neighbors per query \\
        12 & 3, 10, $\uparrow$ & Group & $[256 \times 10 \times 3]$ & Grouped neighborhoods \\
        13 & $\uparrow$ & Neighborhood normalization & $[256 \times 10 \times 3]$ & Translation normalization towards query point \\
        14 & 9, 11 & Group & $[256 \times 10 \times 128]$ & Grouped features \\
        15 & 13, $\uparrow$ & Concat features & $[256 \times 10 \times 131]$ & Grouped features with $xyz$ \\
        16 & $\uparrow$ & (Conv+LeakyReLU) $\times\!2$ & $[256 \times 10 \times 128]$ & Kernel size $1\!\times\!1$, stride 1 \\
        17 & $\uparrow$ & Conv+LeakyReLU & $[256 \times 10 \times 256]$ & Kernel size $1\!\times\!1$, stride 1 \\
        18 & $\uparrow$ & ReduceMax & $[256 \times 256]$ & Maximum over neighborhood features $\rightarrow$ latent representation~$z$\\
        \midrule
        \multicolumn{5}{c}{\textbf{Classifier / Adversary}} \\
        \midrule
        19 & $\uparrow$ & Dense+LeakyReLU & $[256 \times 128]$ & \\
        20 & $\uparrow$ & Dropout & $[256 \times 128]$ & Dropout ratio 50\% \\
        21 & $\uparrow$ & Dense & $[256 \times U_{C,A}]$ & Output logits vector $y_{C,A}$ \\
        22 & $\uparrow$ & Softmax & $[256 \times U_{C,A}]$ & Output probability vector $p_{C,A}$ \\
        \bottomrule
    \end{tabularx}

\end{table*}

\section{Up-Sampling Models}
\label{app:appendix2}

This section gives additional details on the up-sampling experiments for metric verification of Sec.~4.4 in the main paper.
The up-sampling process is based on cylindrical depth projections of the LiDAR point clouds.
Only the vertical resolution of the LiDAR images is enhanced.
The bilinear interpolation is a traditional approach for which we directly used the resize method from TensorFlow (\texttt{tf.image.resize(images, size, method=ResizeMethod.BILINEAR)}).
For all other experiments, we used the generator from the SRGAN architecture~\citep{Ledig2017CVPR} and for the GAN experiments, also the discriminator architecture.
After being processed by the super-resolution networks, the generated point clouds are converted back into lists of points and are fed to the metric network for realism judgement.

\begin{table*}
    \centering

    \caption{
        \textbf{SRGAN Generator Architecture}:
        Detailed network architecture and input format definition of the SRGAN generator~\citep{Ledig2017CVPR}.
        The ID of each row is used to reference the output of the row.
        $\uparrow$ indicates that the layer directly above is an input.
        $N$ denotes the number of measured LiDAR points.
        $H$ denotes the number of layers in the LiDAR sensor and $W$ are the number of layer pulses fired per $360^\circ$ revolution.
        The cylindrical depth projection is either retrieved directly from the raw image of the sensor or with a back-projection by computing $(r,\varphi,\theta)$ from $(x,y,z)$.
        Missing measurements are set to a constant distance in the dense projection and are masked in the loss computation.
    }
    \label{tab:implementation_eval_generator}

    \begin{tabularx}{\linewidth}{rlll X}
        \toprule
        \textbf{ID} & \textbf{Inputs} & \textbf{Operation} & \textbf{Output Shape} & \textbf{Description} \\
        \midrule
        \multicolumn{5}{c}{\textbf{Input features from LiDAR scan}} \\
        \midrule
        1 & LiDAR & $x$, $y$, $z$ & $[N \times 3]$ & Position of each point relative to sensor origin \\
        2 & $\uparrow$ & Projection $(x,y,z) \rightarrow (r,\varphi,\theta)$ & $[H,W,1]$ & Cylindrical depth projection $r$ with $\theta$ over $H$ and $\varphi$ over $W$ \\
        \midrule
        \multicolumn{5}{c}{\textbf{Residual blocks}} \\
        \midrule
        3 & $\uparrow$ & Conv+ParametricReLU & $[H,W,64]$ & Kernel size $9\!\times\!9$, stride 1 \\
        4 & $\uparrow$ & Conv+BN+ParametricReLU & $[H,W,64]$ & Kernel size $3\!\times\!3$, stride 1 \\
        5 & $\uparrow$ & Conv+BN & $[H,W,64]$ & Kernel size $3\!\times\!3$, stride 1 \\
        6 & $\uparrow$, 3 & Add & $[H,W,64]$ & Element-wise addition \\
        7 & $\uparrow$ & Repeat steps (4-6) & $[H,W,64]$ & $\times\!16$ repetition of residual blocks \\
        8 & $\uparrow$ & Conv+BN & $[H,W,64]$ & Kernel size $3\!\times\!3$, stride 1 \\
        9 & $\uparrow$, 3 & Add & $[H,W,64]$ & Element-wise addition \\
        \midrule
        \multicolumn{5}{c}{\textbf{Super-resolution blocks}} \\
        \midrule
        10 & $\uparrow$ & Conv & $[H,W,256]$ & Kernel size $3\!\times\!3$, stride 1 \\
        11 & $\uparrow$ & SubpixelShuffle & $[2 \cdot H,W,128]$ & Reshape by moving values from the channel dimension to the spatial dimension \\
        12 & $\uparrow$ & ParametricReLU & $[2 \cdot H,W,128]$ & \\
        13 & $\uparrow$ & Repeat steps (10-12) & $[f_\text{up} \cdot H,W,128]$ & $\times\!\log_2 f_\text{up}$ repetition with $f_\text{up}$ being the desired up-sampling factor, \ie $f_\text{up}=\{2,4,8\}$ \\
        14 & $\uparrow$ & Conv & $[f_\text{up} \cdot H,W,1]$ & Kernel size $9\!\times\!9$, stride 1 \\
        \bottomrule
    \end{tabularx}

\end{table*}

\tabref{tab:implementation_eval_generator}~lists all layers, inputs, and operations of the SRGAN generator architecture.
In the $\mathcal{L}_{\{1,2\}}$-CNN trainings, a weighted $\mathcal{L}_\alpha$ loss is minimized.
The objective is formulated as
\begin{equation*}
    \min_{\theta_G} \mathcal{L}_\alpha = \min_{\theta_G} \frac{1}{\alpha |\gamma|} \sum_{(i,j)\in\gamma} \left| r_{i,j}^\text{gt} - r_{i,j}^\text{hr} \right|
\end{equation*}
with the set of measured points $\gamma$, and $r^\text{gt}$ being the high-resolution Ground Truth target and $r^\text{hr}$ the prediction
\begin{equation*}
    r^\text{hr} = G_{\theta_G} \left( r^\text{lr} \right)
\end{equation*}
from the low-resolution input $r^\text{lr}$.

\begin{table*}
    \centering

    \caption{
        \textbf{SRGAN Discriminator Architecture}:
        Detailed network architecture and input format definition of the SRGAN discriminator~\citep{Ledig2017CVPR}.
        The input to the network is either the ground truth $r^\text{gt}$ or the prediction from the generator $r^\text{hr}$.
    }
    \label{tab:implementation_eval_discriminator}

    \begin{tabularx}{\linewidth}{rlll X}
        \toprule
        \textbf{ID} & \textbf{Inputs} & \textbf{Operation} & \textbf{Output Shape} & \textbf{Description} \\
        \midrule
        1 & LiDAR & $r^\text{gt}$ or $r^\text{hr}$ & $[f_\text{up} \cdot H,W,1]$ & High-resolution cylindrical depth projection \\
        \midrule
        \multicolumn{5}{c}{\textbf{Conv blocks}} \\
        \midrule
        2 & $\uparrow$ & Conv+LeakyReLU & $[f_\text{up} \cdot H,W,64]$ & Kernel size $3\!\times\!3$, stride 1 \\
        3 & $\uparrow$ & Conv+BN+LeakyReLU & $[\frac{f_\text{up}}{2} H,\frac{1}{4} W,64]$ & Kernel size $5\!\times\!5$, strides $2\!\times\!4$ \\
        4 & $\uparrow$ & Conv+BN+LeakyReLU & $[\frac{f_\text{up}}{2} H,\frac{1}{4} W,128]$ & Kernel size $3\!\times\!3$, stride 1 \\
        5 & $\uparrow$ & Conv+BN+LeakyReLU & $[\frac{f_\text{up}}{4} H,\frac{1}{8} W,128]$ & Kernel size $3\!\times\!3$, stride 2 \\
        6 & $\uparrow$ & Conv+BN+LeakyReLU & $[\frac{f_\text{up}}{4} H,\frac{1}{8} W,256]$ & Kernel size $3\!\times\!3$, stride 1 \\
        7 & $\uparrow$ & Conv+BN+LeakyReLU & $[\frac{f_\text{up}}{4} H,\frac{1}{16} W,256]$ & Kernel size $3\!\times\!3$, strides $1\!\times\!2$ \\
        8 & $\uparrow$ & Conv+BN+LeakyReLU & $[\frac{f_\text{up}}{4} H,\frac{1}{16} W,512]$ & Kernel size $3\!\times\!3$, stride 1 \\
        9 & $\uparrow$ & Conv+BN+LeakyReLU & $[\frac{f_\text{up}}{8} H,\frac{1}{32} W,512]$ & Kernel size $3\!\times\!3$, stride 2 \\
        \midrule
        \multicolumn{5}{c}{\textbf{Reduction}} \\
        \midrule
        10 & $\uparrow$ & Flatten & $[\frac{f_\text{up}}{2} \cdot H \cdot W]$ & \\
        11 & $\uparrow$ & Dense+LeakyReLU & $[1024]$ & \\
        12 & $\uparrow$ & Dense & $[1]$ & \\
        \bottomrule
    \end{tabularx}

\end{table*}

\tabref{tab:implementation_eval_discriminator}~lists all layers, inputs, and operations of the SRGAN discriminator architecture.
Here, an adversarial loss, defined as
\begin{equation*}
    \min_{\theta_G} \max_{\theta_D}
    \left\{
    \log{\left[ D_{\theta_D} \left( r^\text{gt} \right) \right]} +
    \log{\left[ 1 - D_{\theta_D} \left( G_{\theta_G} \left( r^\text{lr} \right) \right) \right]}
    \right\}
\end{equation*}
is minimized.
The Adam optimizer is used for optimization with an initial learning rate of $1e^{-3}$.

\section{Segmentation}
\label{app:appendix3}

For both segmentation models, DarkNet21 and SqueezeSegV2, we use the original PyTorch implementation of Milioto\etal~\citep{Milioto2019IROS}\footnote{Code: \url{https://github.com/PRBonn/lidar-bonnetal}}.
For our experiments, we modified the class labels to have the same classes for all used datasets.
\tabref{tab:appendix_label_mapping}~shows the label mapping from the original dataset to our custom label set for all three datasets.

\begin{table*}
    \centering
    \caption{
        \textbf{Class Mapping}:
        This table shows the detailed class label mapping of the original dataset label ids to our custom mapping used for the segmentation experiments.
    }
    \label{tab:appendix_label_mapping}
    {\small
    \begin{tabularx}{\linewidth}{@{\hskip 2pt} X p{4cm} p{4cm} p{4cm} @{\hskip 2pt}}
        \toprule
        Learned Classes & KITTI & nuScenes & CARLA \\
        \midrule
        unlabeled (0) &
        \makecell[tl]{unlabeled~(0)\\ outlier~(1)\\ on-rails~(16, 256)\\ other-vehicle~(20, 259)\\ other-structure~(52)\\ other-object~(99)} &
        \makecell[tl]{noise~(0)\\ animal~(1)\\ personal-mobility~(5)\\ stroller~(7)\\ wheelchair~(8)\\ barrier~(9)\\ debris~(10)\\ pushable-pullable~(11)\\ trafficcone~(12)\\ bicycle-rack~(13)\\ ambulance~(19)\\ police~(20)\\ trailer~(22)\\ other~(29)\\ ego-vehicle~(31)} &
        \makecell[tl]{unlabeled~(0)\\ other~(3)} \\
        \hline
        person (1) &
        \makecell[tl]{person~(30, 254)\\ bicyclist~(31, 253)\\ motorcyclist~(32, 255)} &
        \makecell[tl]{adult~(2)\\ child~(3)\\ construction-worker~(4)\\ police-officer~(6)} &
        \makecell[tl]{pedestrian~(4)\\ rider~(13)} \\
        \hline
        two-wheeler (2) &
        \makecell[tl]{bicycle~(11)\\ motorcycle~(15)} &
        \makecell[tl]{bicycle~(14)\\ motorcycle~(21)} &
        two-wheeler~(14) \\
        \hline
        large-vehicle (3) &
        \makecell[tl]{bus~(13, 257)\\ truck~(18, 258)} &
        \makecell[tl]{bus~(15, 16)\\ construction vehicle~(18)\\ truck~(23)} &
        - \\
        \hline
        vehicle (4) &
        car~(10, 252) &
        car~(17) &
        car~(10) \\
        \hline
        road (5) &
        \makecell[tl]{road~(40)\\ parking~(44)\\ other-ground~(49)\\ lane-marking~(60)} &
        \makecell[tl]{driveable-surface~(24)\\ flat-other~(25)} &
        \makecell[tl]{road-line~(6)\\ road~(7)} \\
        \hline
        sidewalk (6) &
        sidewalk~(48) &
        sidewalk~(26) &
        sidewalk~(8) \\
        \hline
        terrain (7) &
        terrain~(72) &
        terrain~(27) &
        terrain~(15) \\
        \hline
        construction (8) &
        \makecell[tl]{building~(50)\\ fence~(51)\\ pole~(80)\\ traffic-sign~(81)} &
        manmade~(28) &
        \makecell[tl]{building~(1)\\ fence~(2)\\ pole~(5)\\ wall~(11)\\ traffic-sign~(12)}\\
        \hline
        vegetation (9) &
        \makecell[tl]{vegetation~(70)\\ trunk~(71)} &
        vegetation~(30) &
        vegetation~(9) \\
        \bottomrule
    \end{tabularx}
    }
\end{table*}

\section{Additional Results}
\label{app:appendix4}

Additionally to~\figref{fig:experiments_metric_training_results}, \tabref{tab:experiments_metric_over_classes}~provides class-wise metric results for the datasets with semantic labels.

\begin{table}[h]
    \centering
        \caption{
        \textbf{Class-wise Metric Results}:
        Shown are the class-wise averages of the metric output~$S$ for \Real{}/\Misc{}/\Syn{}.
        In line ``total'', also scores with unknown semantic labels are included.
    }
    \label{tab:experiments_metric_over_classes}
    \begin{tabularx}{\linewidth}{@{\hskip 2pt}X ccc@{\hskip 2pt}}
        \toprule
                        & KITTI       & nuScenes    & CARLA       \\
        \midrule
        person          & .98/.00/.02 & .75/.00/.25 & .14/.00/.86 \\
        two-wheeler     & .96/.00/.04 & .63/.00/.37 & .05/.00/.95 \\
        large-vehicle   &  --/ --/ -- & .68/.00/.31 &  --/ --/ -- \\
        vehicle         & .92/.00/.08 & .81/.00/.19 & .07/.00/.93 \\
        road            & .91/.00/.09 & .85/.00/.15 & .02/.00/.98 \\
        sidewalk        & .88/.00/.12 & .85/.00/.14 & .08/.00/.92 \\
        terrain         & .93/.03/.05 & .77/.04/.19 & .36/.00/.64 \\
        construction    & .62/.00/.38 & .76/.01/.22 & .06/.00/.94 \\
        vegetation      & .87/.00/.13 & .82/.07/.11 & .05/.00/.95 \\
        \midrule
        total           & .80/.00/.20 & .85/.02/.13 & .16/.00/.84 \\
        \bottomrule
    \end{tabularx}
\end{table}

\end{appendices} 

\end{document}